\documentclass[times,review,10pt]{elsarticle}
\usepackage{amssymb}
\usepackage{amsmath}
\usepackage{caption}
\usepackage{array}
\usepackage{stfloats}
\usepackage{url}
\usepackage{verbatim}
\usepackage{graphicx}
\usepackage{booktabs}
\usepackage[table,xcdraw]{xcolor}
\usepackage{multirow}
\usepackage{CJK}
\usepackage{tabularx}
\usepackage{threeparttable}
\usepackage{pifont}
\usepackage{algorithm}
\usepackage{algorithmicx}
\usepackage{algpseudocode}
\usepackage{subcaption}
\usepackage{listings}
\definecolor{codegreen}{rgb}{0,0.6,0}
\definecolor{codegray}{rgb}{0.5,0.5,0.5}
\definecolor{codepurple}{rgb}{0.58,0,0.82}
\definecolor{backcolour}{rgb}{0.95,0.95,0.92}

\lstdefinestyle{mystyle}{
    backgroundcolor=\color{backcolour},   
    commentstyle=\color{codegreen},
    keywordstyle=\color{magenta},
    numberstyle=\tiny\color{codegray},
    stringstyle=\color{codepurple},
    basicstyle=\ttfamily\scriptsize, 
    breakatwhitespace=false,         
    breaklines=true,                 
    captionpos=b,                    
    keepspaces=true,                 
    numbers=left,                    
    numbersep=5pt,                  
    showspaces=false,                
    showstringspaces=false,
    showtabs=false,                  
    tabsize=2,
    language=Python
}
\journal{Pattern Recognition}
\begin{document}
\begin{frontmatter}



\title{EgoHieraLoc: A Cortically Inspired Hierarchical Segmentation-Guided Framework for Egocentric Visual Query Localization}

\author[label1]{Yifei Cao}
\author[label2]{Guolong Wang}
\author[label3]{Mingliang Hou}
\author[label1]{Xiya Bu}
\author[label1]{Daming Liu}
\author[label1]{Yu Liu\corref{cor1}}

\cortext[cor1]{Corresponding author: Yu Liu (e-mail: yuliu@dlut.edu.cn)}

\affiliation[label1]{organization={Dalian University of Technology},
             city={Dalian},
             country={China}}

\affiliation[label2]{organization={University of International Business and Economics},
             city={Beijing},
             country={China}}

\affiliation[label3]{organization={Jinan University},
             city={Guangzhou},
             country={China}}
\begin{abstract}
Visual query localization (VQL) aims to retrieve and re-localize a queried object in egocentric videos, yet remains challenging when object boundaries are ambiguous and global context cannot effectively guide fine-grained localization. Human vision handles such ambiguity through a hierarchical process: it rapidly screens foreground candidates, selectively attends to the target despite distractors, refines perception via feedback between global context and local detail, and, when a single view is unreliable, integrates evidence across viewpoints according to its credibility. Inspired by these competencies, we propose \textbf{EgoHieraLoc}, a unified framework for VQL-2D and VQL-3D. A Discriminative Parsing Module first extracts foreground-aware query representations using segmentation priors; a Query-Aware Module then performs robust target localization through discriminative correlation filtering with deformable modeling; and a Regional Adaptation Module feeds multi-scale context back into local regions to recover precise object boundaries. To extend this perceptual hierarchy to 3D localization, we introduce Geometric-Semantic Joint Confidence (GSJC), which multiplicatively couples segmentation confidence with local depth consistency, multi-view back-projection consistency, and triangulation-baseline quality, so that a viewpoint contributes to the 3D estimate only when it is credible both semantically and geometrically. Extensive experiments demonstrate state-of-the-art performance on both VQL-2D and -3D benchmarks.
\end{abstract}


\end{frontmatter}



\section{Introduction}
\begin{figure*}[tp]
\centering
\includegraphics[width=\textwidth]{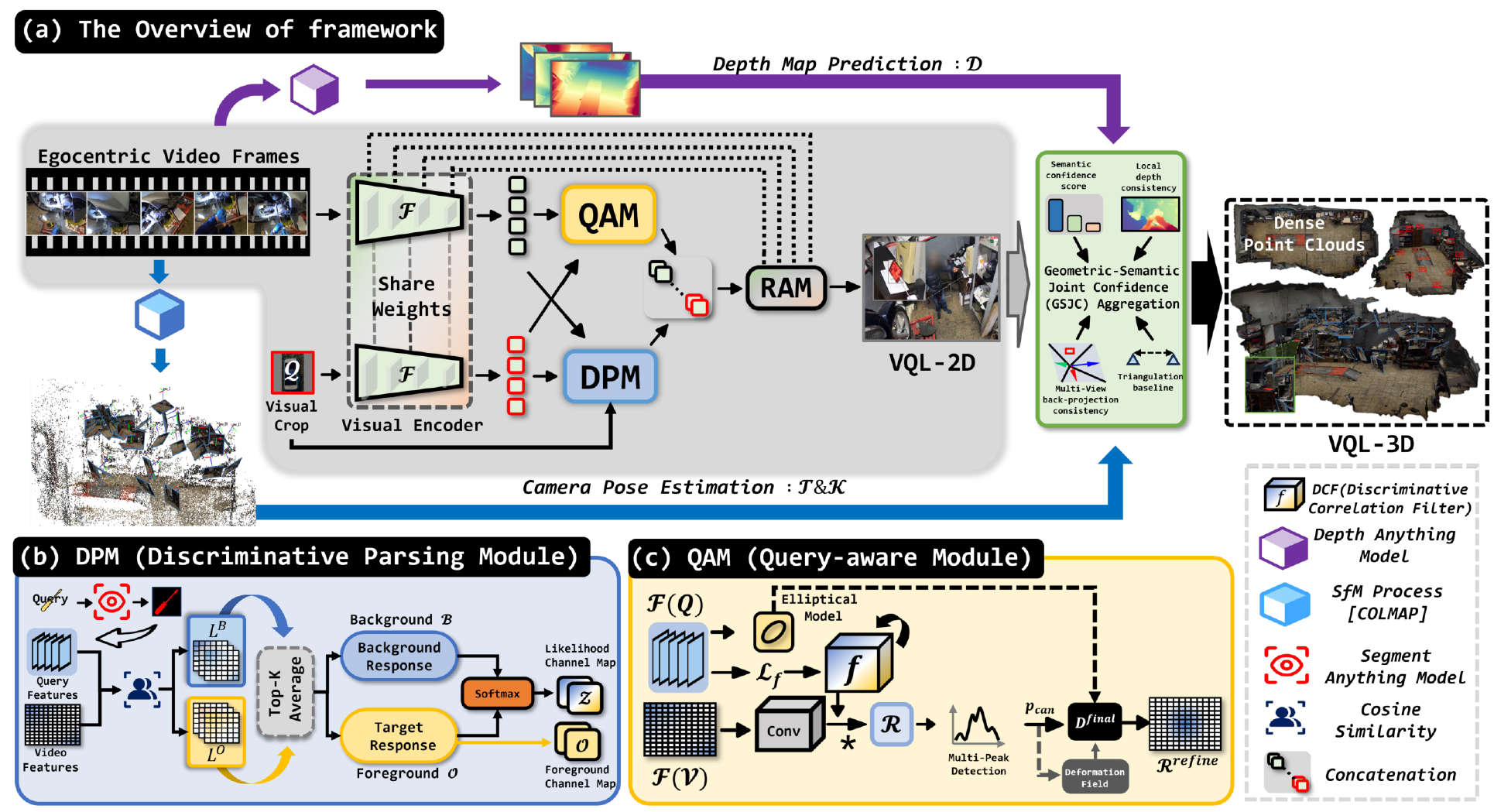}
\caption{\textbf{The framework of the EgoHieraLoc}. (a) represents the overall architecture of EgoHieraLoc, with the content of the dashed box indicating the 2D task branch. This branch consists of three key components: the Discriminative Parsing Module (b, blue block), the Query-Aware Module (QAM) (c, yellow block), and the Regional Adaptation Module (RAM) (see Fig.\ref{fig:ram}). These modules collaboratively establish semantic associations between the target and its surrounding environment in the 2D space to enable effective target retrieval. The remaining part corresponds to the 3D process, which integrates 2D predictions, depth estimation, and camera poses, and employs GSJC to generate the final 3D results.}
\label{fig:overall}
\end{figure*}

Visual query localization (VQL), a fundamental yet challenging task in computer vision, has recently attracted significant attention for its central role in embodied AI, robotics, and AR/VR~\cite{egooutlook,ego4d,relocate,eagle}. VQL enables an agent to retrieve and re-localize a previously seen object in egocentric videos, formulated as recovering the object's last appearance as a 2D bounding box or a 3D vector. Existing methods either couple detection and tracking modules~\cite{ego4d,eivul,egocol,wallet,relocate,egoloc}, or reduce the difficulty of spatio-temporal matching by explicitly modeling inter- and intra-frame relations~\cite{vqloc,prvql,herovql,biovql}. However, they often assume that the query object exhibits clearly visible boundaries in the reference frame, an assumption that frequently fails under egocentric motion blur, occlusion, and drastic scale change, making precise localization hard to sustain during retrieval. Moreover, current methods still struggle with fine-grained boundary segmentation and show limited ability to relate global context to local appearance. These issues collectively cap further progress in VQL.

Humans, in contrast, re-localize objects in dynamic environments with remarkable robustness, owing to the hierarchical and feedback-driven organization of the visual system. Perception begins with rapid foreground screening that segregates candidate objects from cluttered surroundings; attention is then selectively steered toward the target of interest despite distractors and appearance changes; and higher cortical areas refine this percept through top-down feedback that reconciles global context with local detail. Crucially, when a single observation is ambiguous or unreliable, humans do not weigh all evidence equally, but integrate cues across viewpoints according to their perceived reliability to form a stable spatial estimate. This progression, from coarse screening, to attentive localization, to feedback-based refinement, and finally to reliability-weighted spatial integration, provides valuable inspiration for machine vision.

Motivated by this, we propose EgoHieraLoc, a human-vision-inspired framework that instantiates these four competencies as three cooperating 2D modules and a geometry-aware 3D fusion scheme, forming a perceptual flow from semantic guidance to geometric refinement, and endowing the model with both global understanding and precise localization in complex videos. Concretely, the \textit{Discriminative Parsing Module} (DPM) realizes rapid foreground screening: it exploits the strong priors of the Segment Anything Model (SAM)~\cite{sam} to generate fine-grained query representations and separate the target from background clutter within the search region. The \textit{Query-aware Module} (QAM) implements attentive localization: it focuses on the query region with a discriminative correlation filter (DCF)~\cite{dcf}, strengthening resistance to distractors and robustness to scale variation through dynamic deformable modeling. The \textit{Regional Adaptation Module} (RAM) performs feedback-based refinement: simulating the re-entry and integration of higher cortical signals, it aggregates multi-scale backbone features to refine and restore the final segmentation, implicitly associating global context with local appearance. Each module plays a distinct role and complements the others, together forming a powerful and robust pipeline.

The same principle extends naturally from a single image to 3D space. Once 2D predictions are lifted into 3D by back-projection through SfM poses and robust monocular depth, a stable estimate must be recovered from many noisy per-frame observations, yet not all views deserve equal trust: a confident mask can still yield an unreliable 3D point when its depth is locally ambiguous, when it disagrees with the multi-view consensus, or when its viewing baseline is near-degenerate. Mirroring how humans weight evidence by its reliability, we propose a novel \textit{Geometric-Semantic Joint Confidence (GSJC)} weighting scheme that assigns each view a credibility by multiplicatively coupling its segmentation confidence with three geometric factors: local depth consistency, multi-view reprojection consistency, and triangulation-baseline quality. Because these factors are combined multiplicatively, a view contributes only when it is credible along every axis at once, so any single unreliable cue suppresses its influence, while a small clamp keeps the confidence-weighted aggregation well-defined and lets it gracefully reduce to a uniform average when all views are unreliable. GSJC thus turns multi-view fusion from a heuristic average into a reliability-driven decision, yielding stable 3D displacement estimates and providing a principled solution for both VQL-2D and VQL-3D within a single framework. Our method achieves state-of-the-art performance on all VQL tasks, demonstrating the potential of this framework as a strong new baseline.

Our main contributions are summarized as follows:
\begin{itemize}
    \item We propose EgoHieraLoc, a novel human-vision-inspired framework that unifies VQL-2D and VQL-3D through three specialized, synergistic modules: a Discriminative Parsing Module (DPM) for fine-grained query representation, a Query-aware Module (QAM) for robust target localization, and a Regional Adaptation Module (RAM) for high-fidelity segmentation restoration, jointly emulating the hierarchical mechanism of the human visual system.
    \item We present the Geometric-Semantic Joint Confidence (GSJC) weighting scheme, which multiplicatively fuses segmentation confidence with local depth consistency, multi-view reprojection consistency, and triangulation-baseline quality, enabling reliability-driven multi-view aggregation and stable 3D displacement estimation.
    \item Extensive experiments show that EgoHieraLoc achieves superior performance on both VQL-2D and VQL-3D benchmarks, demonstrating the framework's potential as a unified solution and robust baseline for VQL in egocentric vision.
\end{itemize}
\section{Related Works}
\label{sec:relatedwork}
\noindent\textbf{Few-shot Visual Object Tracking}. VQL links closely to few-shot visual object tracking, as both require localizing targets under limited prior knowledge. Traditional short- and long-term trackers \cite{ego4d,isfirst,egovot,zsl,srrt,lgtrack} rely on similarity matching or discriminative modeling. Siamese trackers \cite{siamfc,siamrpn,siamrpn++,siamfcplus} measure similarity via dual-branch feature alignment, emphasizing global matching and generalization. Conversely, discriminative methods stabilize tracking by decoupling it into classification and regression. Recent work shifts toward few-shot video segmentation and instance-level tracking: \cite{siammask} integrates saliency detection; \cite{stark,seqformer} use Transformers for spatio-temporal dependencies; \cite{d3s,masktrack,feelvos,vistr} extend instance segmentation to videos; and \cite{sam,sam2} show zero-shot potential. While the Ego4D-VQ baseline uses tracking-centric Siam-RCNN \cite{siamrcnn}, such methods typically assume canonically cropped templates. In real-world VQL, queries often involve occlusions or motion blur, degrading performance when appearance discrepancies exist between query and search domains. We propose a segmentation-centric path for visual query modeling. By generating pixel-level boundary-aware representations, our approach captures robust appearance features for stable spatio-temporal localization.

\noindent\textbf{Visual Query Localization with 2D\&3D}. VQL comprises 2D (VQL-2D) and 3D (VQL-3D) tasks. VQL-2D localizes the target's spatial-temporal occurrence, while VQL-3D pinpoints its 3D location. Current research predominantly focuses on the 2D task. Ego4D\cite{ego4d,eivul} proposed a three-stage approach integrating frame-wise detection and bidirectional tracking. CocoFormer\cite{vq2d,wallet} used negative frames and optimized proposal generation to improve VQL-2D performance. VQLoc\cite{vqloc} simplified architectures by implementing end-to-end position prediction with a Transformer. HERO-VQL\cite{herovql} combines TAG and Egocentric Augmentation-based Consistency Training (EgoACT) to tackle challenges like drastic viewpoint changes and occlusions. RELOCATE\cite{relocate} achieves efficient localization in long videos by using region-level features from pre-trained models combined with bidirectional tracking. PRVQL\cite{prvql} progressively extracts target-relevant appearance and spatial knowledge from the video to refine both query and video features. For VQL-3D, EgoCOL\cite{egocol} used sparse camera reconstruction in a two-fold manner, video and scan independently, to estimate the camera pose of egocentric frames in 3D renders with high recall and precision. EgoLoc\cite{egoloc} entangles 3D multi-view geometry with 2D object retrieval to improve 3D localization performance. Despite progress on each task in isolation, the 2D and 3D pipelines remain largely disjoint: the former concentrates on semantic matching, while the latter relies on geometric reconstruction, without a shared representation that allows semantic evidence to modulate geometric fusion. Inspired by the hierarchical organization of human vision, our framework couples the two within a single pipeline: the 2D branch supplies both the localization and a per-view semantic confidence, which is combined with three geometric reliability factors, namely local depth consistency, multi-view reprojection consistency, and triangulation-baseline quality, through the proposed Geometric-Semantic Joint Confidence (GSJC) aggregation. We note that this coupling is \emph{directional}, propagating from 2D semantics to 3D geometry rather than forming a mutual refinement loop; such a design keeps the input--output contract of each stage interpretable while still letting mask reliability determine which viewpoints dominate the final 3D estimate.

\noindent\textbf{Egocentric 3D Visual Understanding}. Egocentric 3D Visual Understanding enables first-person perception for AR/VR and robotic navigation. Structure-from-Motion (SfM)\cite{sfm} provides the foundational principles for recovering 3D scene structure and camera poses from multiple 2D views. However, applying classical SfM to egocentric videos presents unique challenges, primarily due to drastic viewpoint changes, severe motion blur, and the inherent dynamic nature of human-centric interactions. Moreover, many downstream tasks, such as precise object localization, do not necessitate a full, dense 3D reconstruction of the entire scene. To address these challenges, various advanced methods have emerged. Some approaches approximate Simultaneous Localization and Mapping (SLAM) as an SfM estimation over a sliding window for enhanced robustness in egocentric settings, such as EgoSLAM\cite{egoslam}. Others focus on dense scene representation and reconstruction: NeuralDiff\cite{neuraldiff} reconstructs dynamic video scenes using Neural Radiance Fields (NeRF)\cite{nerf}, while EgoLifter\cite{egolifter} integrates instance-level segmentation via SAM to refine object geometry. More recently, methods like DUST3R\cite{dust3r} achieve unconstrained 3D scene reconstruction, and EgoGaussian\cite{egogaussian} utilizes Gaussian Splatting\cite{3dgs} for explicit object-background separation. While these methods significantly advance 3D understanding through comprehensive 3D models, they primarily target full scene reconstruction and offer limited mechanisms for leveraging 2D semantic cues to enhance precise target localization, especially when reconstruction is computationally expensive or unnecessary. Our work departs from complete 3D scene reconstruction, instead focusing on robust 3D localization of targets in egocentric sequences by exploiting 2D information. We propose a multi-view aggregation function that fuses 3D location estimations across frames, weighted by a novel Geometric-Semantic Joint Confidence (GSJC) score. The GSJC scheme combines semantic segmentation confidence with three geometric reliability factors: depth uncertainty, multi-view reprojection consistency, and triangulation baseline quality. This semantic-geometric synergy effectively filters low-confidence 3D predictions and prioritizes reliable cues across viewpoints, significantly improving accuracy and robustness of 3D displacement estimation and target localization in complex, dynamic egocentric environments.
\begin{figure*}[tp]
\centering
\includegraphics[width=\textwidth]{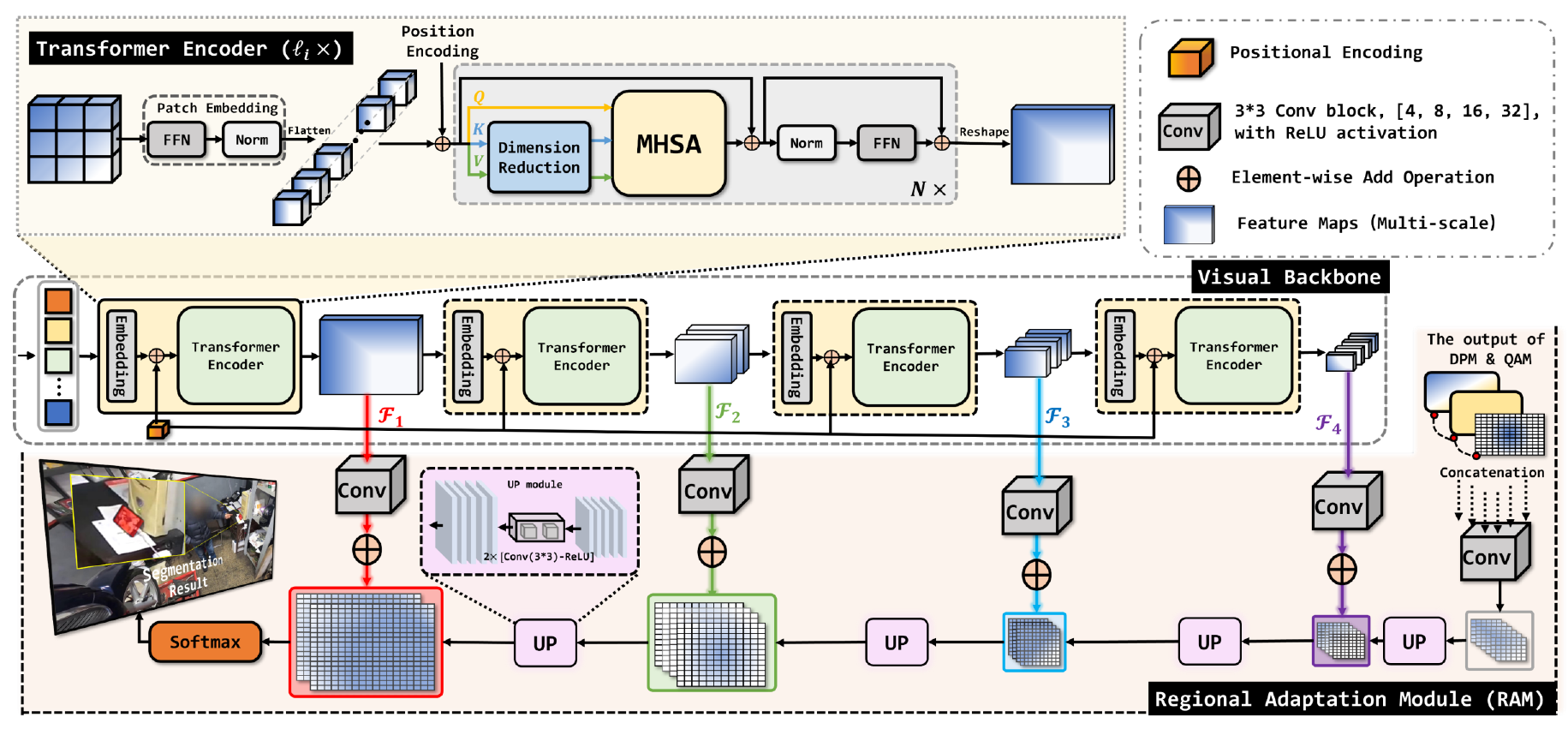}
\caption{\textbf{The architecture of RAM}. The backbone adopts a pyramidal structure, containing four stages, each integrating an encoder module and progressively downsampling to reduce resolution via a dimensional collapse module (DCM). The RAM (bottom dashed box) receives the cascaded DPM and QAM outputs, fuses multi-level features through convolution and upsampling modules, and gradually restores the output to the original resolution.}
\label{fig:ram}
\end{figure*}
\section{The Proposed Method}
\label{sec:method}
\subsection{Task Formulation}
\label{app:task}
Given a visual query $\mathcal{Q}$ (an image crop) and an egocentric video sequence $\mathcal{V}=\{v_i\}_{i=0}^{N_V-1}$ with $N_V$ frames, our framework addresses two complementary tasks. The VQL-2D task predicts a spatio-temporal response track $\mathcal{R}=\{b_t\}_{t=s}^e$, where $b_t=[x_t,y_t,w_t,h_t]$ denotes the 2D bounding box centered on the target in frame $v_t$, with $s$ and $e$ representing the start and end frame indices. Building on these 2D localization results, the VQL-3D task determines the 3D position of the target relative to the camera center of the query frame. We first apply SfM\cite{sfm} and depth estimation\cite{depth} to compute the camera poses $\{\mathcal{T}_{v_i}\}_{i=0}^{N_V-1}$ and depth maps $\{\mathcal{D}_{v_i}\}_{i=0}^{N_V-1}$ for all video frames. We then combine the 2D predictions with the estimated camera poses, depth maps, and a multi-view aggregation function to recover the 3D position $[\hat{x}, \hat{y}, \hat{z}]$ of the object in the world coordinate system. Finally, we compute the relative displacement offset $\delta$ by projecting $\mathcal{T}_{v_i}$ relative to the current frame.
\subsection{Discriminative Parsing Module (DPM)}
To simulate the process of the human visual cortex rapidly screening targets, we utilize a strong-prior SAM\cite{sam}\footnote{Details of the degradation strategy for SAM failures are provided in \ref{app:qamfailure}.} to generate query representations to distinguish the target from the background within the search area. Specifically, SAM segments query $\mathcal{Q}\in \mathbb{R}^{h\times w \times c}$, producing binary mask $\mathcal{S(\mathcal{Q})}\in\{0,1\}^{h\times w}$. Query features $\mathcal{F}(\mathcal{Q})\in\mathbb{R}^{h'\times w'\times d}$, extracted by the backbone, are spatially aligned with the mask via bilinear interpolation. We employ a scaling ratio $s=\lfloor h/h' \rfloor$ to map feature coordinates $(x',y')$ to $(x's,y's)$. This yields mask-guided object and background feature sets, both derived from $\mathcal{Q}$: 
\begin{equation}
\begin{cases}
    \mathcal{M}_O = \{\mathcal{F}(\mathcal{Q})_{(x',y')} \mid \mathcal{S}(\mathcal{Q})_{(x's,y's)}=1\},\\\mathcal{M}_B = \{\mathcal{F}(\mathcal{Q})_{(x',y')} \mid \mathcal{S}(\mathcal{Q})_{(x's,y's)}=0\}.
\end{cases}
\end{equation}
Each set $\mathcal{M}_O$ (or $\mathcal{M}_B$) contains multiple $d$-dimensional feature vectors corresponding to the foreground (or background) pixels of the query. Search region features $\mathcal{F}(v_i)\in\mathbb{R}^{H\times W\times D}$ are processed by two simple CBN layers (Conv-BN-ReLU). $CBN_1$ performs $1 \times 1$ channel-wise linear transformations, focusing on local, fine-grained features, while $CBN_2$ utilizes a $3\times3$ convolution to capture spatial contextual relationships. The formulation is as follows: 
$\mathcal{F}(v_i)' = CBN_2(CBN_1(\mathcal{F}(v_i))) \in \mathbb{R}^{H_{\xi}\times W_{\xi}\times D_{\xi}},$
where $H_{\xi}$ and $W_{\xi}$ are the spatial dimensions after
stride-$\xi$ downsampling, and $D_{\xi}$ is the feature dimension. For each spatial location $(x,y)$ in $\mathcal{F}(v_i)'$, we compute the cosine similarity between $\mathcal{F}(v_i)'_{(x,y)}$ and each individual feature vector $m_o \in \mathcal{M}_O$ and $m_b \in \mathcal{M}_B$. This generates a set of raw foreground matching scores $L_{(x,y)}^O=\{L_{(x,y),m_o}^O|m_o \in \mathcal{M}_O\}$ and background matching scores $L_{(x,y)}^B=\{L_{(x,y),m_b}^B|m_b \in \mathcal{M}_B\}$, as formulated below:
\begin{equation}
    \begin{cases}
       L_{(x,y),m_o}^O =\frac{\mathcal{F}(v_i)'_{(x,y)} \cdot m_o}{\max(\|\mathcal{F}(v_i)'_{(x,y)}\|_2\cdot\|m_o\|_2, \epsilon)},\\L_{(x,y),m_b}^B = \frac{\mathcal{F}(v_i)'_{(x,y)} \cdot m_b}{\max(\|\mathcal{F}(v_i)'_{(x,y)}\|_2\cdot\|m_b\|_2, \epsilon)},
    \end{cases}
\end{equation}
where $\|\cdot\|_2$ denotes the $\ell_2$-norm, and $\epsilon=1e^{-8}$ prevents division-by-zero errors. Top-K average is used to aggregate responses:
\begin{equation}
    \begin{cases}\mathcal{O}= \frac{1}{K_{DPM}} \sum_{L_{(x,y)}^O\in \mathrm{TopK_{DPM}}(L_{(x,y)}^O, K_{DPM})} L_{(x,y)}^O, \\ \mathcal{B}= \frac{1}{K_{DPM}} \sum_{L_{(x,y)}^B\in \mathrm{TopK_{DPM}}(L_{(x,y)}^B, K_{DPM})} L_{(x,y)}^B,
    \end{cases}
\end{equation}
where $K_{DPM}$ is set to 3, and $\mathrm{TopK_{DPM}}(\cdot,K_{DPM})$ selects the $K_{DPM}$ highest scores from the provided set for averaging. Finally, the likelihood-channel map, $\mathcal{Z}$, is obtained via a softmax operation: $\mathcal{Z}= \mathrm{softmax}([\mathcal{O}, \mathcal{B}]).$ Consequently, the foreground map ($\mathcal{O}$) and the likelihood-channel map ($\mathcal{Z}$) serve as outputs of the DPM. 
\subsection{Query-aware Module (QAM)}
\label{sec:qam}
To refine localization and enhance robustness against similar distractors and scale variations, 
we design the Query-aware Module (QAM) by integrating Discriminative Correlation Filters (DCF) 
with multi-peak detection and dynamic elliptical deformable constraints. This module operates 
in a frame-by-frame manner, leveraging temporal information to maintain spatial consistency 
while adapting to object deformations. The DCF $f$ operates on the query features $\mathcal{F}(\mathcal{Q})$ 
and is defined by the following ridge-regression objective, which encourages 
the filter response on the query to match an ideal Gaussian target:
\begin{equation}
\label{eq:dcf_loss}
\mathcal{L}_f = \left\|\mathcal{FT}^{-1}\!\left(\mathcal{FT}(\mathcal{F}(\mathcal{Q})) \odot 
\overline{\mathcal{FT}(f)}\right) - y\right\|^2_2 + \lambda\|f\|^2_2,
\end{equation}
where $\mathcal{FT}(\cdot)$ and $\mathcal{FT}^{-1}(\cdot)$ denote the Fourier transform and 
its inverse, respectively, $y \in \mathbb{R}^{h \times w}$ is an ideal Gaussian response map 
centered at the query, and $\lambda\|f\|^2_2$ is a ridge regularization term. For a fixed feature representation $\mathcal{F}(\mathcal{Q})$, the objective in 
Eq.(\ref{eq:dcf_loss}) is a linear ridge-regression problem that admits an exact 
\emph{closed-form solution} in the frequency domain:
\begin{equation}
\label{eq:dcf_closed_form}
\hat{f}^{*} = \frac{\overline{\mathcal{FT}(\mathcal{F}(\mathcal{Q}))} \odot \mathcal{FT}(y)}
{\mathcal{FT}(\mathcal{F}(\mathcal{Q})) \odot \overline{\mathcal{FT}(\mathcal{F}(\mathcal{Q}))} + \lambda},
\end{equation}
where $\hat{f}^{*}=\mathcal{FT}(f^{*})$ and all operations are element-wise. 
Consequently, the filter $f^{*}$ is never obtained through iterative gradient descent; 
it is computed analytically from the query features via Eq.(\ref{eq:dcf_closed_form}). 
The objective in Eq.(\ref{eq:dcf_loss}) therefore serves a dual role. (i) It provides the closed-form filter $f^{*}$ used for localization. (ii) During end-to-end training, the residual of Eq. (\ref{eq:dcf_loss}) 
(evaluated at the closed-form optimum $f^{*}$) is back-propagated only into the 
shared backbone, shaping $\mathcal{F}(\mathcal{Q})$ so that the analytically computed 
filter yields a sharp, Gaussian-like response. This explains the behavior observed in the 
training dynamics\footnote{Detailed analysis of training dynamics is provided in \ref{app:trainingloss}.}: the DCF loss $\mathcal{L}_f$ decreases over iterations 
not because $f$ is optimized step-by-step, but because the backbone features it 
depends on become progressively more discriminative, allowing the closed-form filter to 
reach a stable attractor. This design follows the differentiable-DCF paradigm 
of modern trackers~\cite{atom,dimp}, where the correlation filter is solved 
in closed form per query while the feature extractor is learned end-to-end. Once the query-specific filter $f^{*}$ is obtained, it is applied to every video frame 
$v_i$ without any further update to extract a response map:
\begin{equation}
\label{eq:response_map}
R_{v_i} = \mathcal{FT}^{-1}\left(\mathcal{FT}(\mathcal{F}(v_i)) \odot \overline{\mathcal{FT}(f^*)}\right),
\end{equation}
where $\mathcal{F}(v_i)$ is the backbone feature of frame $v_i$. The response map 
$R_{v_i} \in \mathbb{R}^{H \times W}$ indicates correlation strength at each spatial location. To handle response maps with multiple local maxima (caused by occlusions, motion blur, or 
background clutter), we apply an adaptive threshold-based binarization rather than single-peak 
selection. First, we filter low-response regions using a statistical threshold $T_{v_i}$:
$T_{v_i} = \mu_{R_{v_i}} + \alpha_{\text{QAM}} \cdot \sigma_{R_{v_i}}$, where $\mu_{R_{v_i}}$ and $\sigma_{R_{v_i}}$ are the mean and standard deviation of 
$R_{v_i}$ respectively, and $\alpha_{\text{QAM}} = 1.0$ is a sensitivity parameter controlling the 
threshold aggressiveness. This produces a binary mask $M_{v_i} \in \{0, 1\}^{H \times W}$. We identify connected regions in $M_{v_i}$ using 8-connectivity 
and extract the peak location from each connected component:
\begin{equation}
\label{eq:candidate_peaks}
p_k^{(v_i)} = \operatorname*{argmax}_{(x,y) \in CC_k^{(v_i)}} R_{v_i}(x, y), \quad k = 1, 2, \ldots, N_{CC},
\end{equation}
where $CC_k^{(v_i)}$ denotes the $k$-th connected component and $N_{CC}$ is the total number 
of components. To reduce noise from spurious background peaks, we retain the Top-$K_{\text{QAM}} = 5$ 
candidates with the highest response values. We then weight these candidates by both their response 
strength and the area of their corresponding connected components:
\begin{equation}
\label{eq:primary_candidate}
p_{\text{can}}^{(v_i)} = \operatorname*{argmax}_{p_k^{(v_i)} \in \text{Top-}K_{\text{QAM}}} 
\left[R_{v_i}(p_k^{(v_i)}) \cdot \mathcal{A}(CC_k^{(v_i)})\right],
\end{equation}
where $\mathcal{A}(CC_k^{(v_i)})$ is the pixel count of the connected component. This 
weighting scheme favors large, high-confidence regions over small noise peaks, improving robustness 
to clutter. At the first frame ($i=0$), we perform a direct initialization without recursive refinement:
\begin{equation}
\label{eq:init_protocol}
\begin{cases}
p_{\text{can}}^{(v_0)} = \operatorname*{argmax}_{(x,y)} R_{v_0}(x, y) \\
p_{\text{mod}}^{(v_0)} = p_{\text{can}}^{(v_0)}  \\
\Delta^{(v_0)} = 0  \\
\Omega^{(v_0)} = 0  \\
a^{(v_0)}, b^{(v_0)} = \frac{\text{width}_{\mathcal{Q}}}{2}, \frac{\text{height}_{\mathcal{Q}}}{2}
\end{cases}
\end{equation}
where $\text{width}_{\mathcal{Q}}$ and $\text{height}_{\mathcal{Q}}$ are computed from the SAM-generated mask fitting. This initialization ensures stable 
bootstrapping without accumulating errors from ill-defined prior frames. For subsequent frames ($i \geq 1$), we employ a two-level temporal smoothing mechanism: instantaneous 
deformation and cumulative deformation. The instantaneous deformation field captures frame-to-frame motion:
\begin{equation}
\label{eq:instantaneous_deformation}
\Delta^{(v_i)} = p_{\text{can}}^{(v_i)} - p_{\text{mod}}^{(v_{i-1})},
\end{equation}
which measures the displacement between the current frame's candidate peak and the 
previous frame's refined position. To prevent temporal jitter and accumulation of noise, we maintain 
a cumulative deformation field $\Omega^{(v_i)}$ that exponentially decays older motion:
\begin{equation}
\label{eq:cumulative_deformation}
\Omega^{(v_i)} = \gamma_{\text{QAM}} \cdot \Omega^{(v_{i-1})} + (1 - \gamma_{\text{QAM}}) \cdot \Delta^{(v_i)},
\end{equation}
where $\gamma_{\text{QAM}} = 0.4$ is the decay factor. This recursive scheme maintains an 
exponential moving average of motion, balancing historical and current information. For the first 
frame, $\Omega^{(v_0)} = 0$. To handle scale variations and non-rigid object deformations in egocentric views, we model the 
query's shape using a dynamic ellipse parameterized by semi-major and semi-minor axes. Let 
$w^{(v_i)}$ and $h^{(v_i)}$ denote the bounding-box width and height of the connected component 
containing the candidate peak:
\begin{equation}
\label{eq:bbox_from_cc}
\begin{cases}
w^{(v_i)} = \max_{x \in CC_{\text{can}}^{(v_i)}} x - \min_{x \in CC_{\text{can}}^{(v_i)}} x, \\
h^{(v_i)} = \max_{y \in CC_{\text{can}}^{(v_i)}} y - \min_{y \in CC_{\text{can}}^{(v_i)}} y,
\end{cases}
\end{equation}
where $CC_{\text{can}}^{(v_i)}$ is the connected component associated with $p_{\text{can}}^{(v_i)}$. 
The ellipse semi-axes are recursively updated with temporal smoothing:
\begin{equation}
\label{eq:ellipse_update}
\begin{cases}
a^{(v_i)} = \eta_{\text{QAM}} \cdot a^{(v_{i-1})} + (1 - \eta_{\text{QAM}}) \cdot \frac{w^{(v_i)}}{2}, \\
b^{(v_i)} = \eta_{\text{QAM}} \cdot b^{(v_{i-1})} + (1 - \eta_{\text{QAM}}) \cdot \frac{h^{(v_i)}}{2},
\end{cases}
\end{equation}
where $\eta_{\text{QAM}} = 0.7$ controls the smoothing strength. The initial parameters 
$a^{(v_0)}$ and $b^{(v_0)}$ are derived directly from the query size. To ensure the standard ellipse form where the first axis is 
the semi-major axis (longer dimension), we enforce $a^{(v_i)} \geq b^{(v_i)}$ via automatic swapping:
\begin{equation}
\label{eq:axis_ordering}
(a^{(v_i)}, b^{(v_i)}) \leftarrow 
\begin{cases}
(a^{(v_i)}, b^{(v_i)}) & \text{if } a^{(v_i)} \geq b^{(v_i)}, \\
(b^{(v_i)}, a^{(v_i)}) & \text{otherwise}.
\end{cases}
\end{equation}
The ellipse is aligned with the image coordinate axes; its orientation angle 
is fixed at $\theta_{ellipse} = 0°$. This alignment choice is motivated by two factors: (i) VQL 
output format specifies axis-aligned bounding boxes, and (ii) in egocentric videos, primary deformations 
are due to perspective scaling rather than in-plane rotation. This simplification is validated by 
ablation studies (Fig. 7), where removing the elliptical constraint degrades performance by 1.19\%. To correct the response map for predicted object motion, we apply a displacement-based shift. For 
each coordinate $i = (x_i, y_i)$ in $R_{v_i}$, we compute the aligned coordinate:
\begin{equation}
\label{eq:motion_compensation}
i' = i + \psi(\Omega^{(v_i)}),
\end{equation}
where $\psi(\cdot)$ is the pixel-rounding function that discretizes the continuous cumulative 
deformation into pixel coordinates: $\psi(\Omega^{(v_i)}) = \left\lfloor \Omega^{(v_i)} + 0.5 \right\rfloor.$ This rounding operation aligns the floating-point motion estimate with the discrete image 
lattice, ensuring numerical stability in subsequent distance computations. We determine the refined query location by minimizing a combined 
distance metric that accounts for both Euclidean proximity and elliptical shape constraints:
\begin{equation}
\label{eq:distance_metric}
\text{Dis}_{i}^{\text{mod},(v_i)} = \|i' - p_{\text{can}}^{(v_i)}\|_2 + \lambda_e \cdot 
\sqrt{\frac{(x_i - p_x)^2}{(a^{(v_i)})^2} + \frac{(y_i - p_y)^2}{(b^{(v_i)})^2}},
\end{equation}
where $(p_x, p_y)$ denotes the coordinates of the candidate peak $p_{\text{can}}^{(v_i)}$, 
and $\lambda_e = 0.5$ weights the elliptical constraint. The first term enforces proximity to the 
motion-compensated candidate, while the second term applies an anisotropic spatial prior that penalizes 
deviations along the principal axes of the ellipse. The refined position is then:
\begin{equation}
\label{eq:refined_position}
p_{\text{mod}}^{(v_i)} = \operatorname*{argmin}_i \text{Dis}_{i}^{\text{mod},(v_i)}.
\end{equation}
This refined position $p_{\text{mod}}^{(v_i)}$ becomes the reference for the next frame's instantaneous 
deformation computation (Eq. (\ref{eq:instantaneous_deformation})). After dimensional alignment with the DPM response map, the QAM location 
response map is concatenated with the DPM foreground likelihood map and fed into the Regional Adaptation 
Module (RAM) for further refinement.
\subsection{Regional Adaptation Module (RAM)}
To better simulate the top-down and bottom-up signal integration of the advanced cortex, and inspired by \cite{pvt}, we aggregate multi-scale features from the backbone network to refine and restore the final segmentation result. This process implicitly models the relationship between global context and local features during restoration. As shown in Fig.\ref{fig:ram}, the hierarchical features $\mathcal{F}_i\in \{\mathcal{F}_1,\mathcal{F}_2,\mathcal{F}_3, \mathcal{F}_4\}$
 at four different scales are generated through four stages, each employing a similar architectural design. Initially, we transform the dimensions of the features from the video frames (which we call the search region). In the first stage, a $1 \times 1$ convolution block is used to reduce the feature channel dimension from $C$ to $D_s$, constructing the embeddings $\mathcal{F}_1 \in \mathbb{R}^{(T \times H_s \times W_s) \times D_s}$. Each subsequent stage takes the feature embeddings from the previous stage as input, generating features $\mathcal{F}_2, \mathcal{F}_3$, and $\mathcal{F}_4$, which correspond to progressively larger strides relative to the input. This demonstrates a hierarchical downsampling strategy. Specifically, we denote the patch size at stage $i$ as $P_i$. At the beginning of stage $i$, the input feature $\mathcal{F}_{i-1} \in \mathbb{R}^{T \times H_{i-1} \times W_{i-1} \times C_{i-1}}$ is evenly divided into $N_P = \frac{T \times H_{i-1} \times W_{i-1}}{P_i^2}$ patches. Each patch is then flattened into a vector of dimension $P_i^2 C_{i-1}$ and projected into an embedding space of $C_i$ dimensions. After the linear projection layer, the shape of the embedded patches can be represented as $(T \times \frac{H_{i-1}}{P_i} \times \frac{W_{i-1}}{P_i}) \times C_{i}$, where the height and width are downsampled by a factor of $P_i$ compared to the input. This process is formalized as follows:
\begin{equation}
\begin{aligned}
    \mathcal{F}_i&=\ell_{i-1}(\text{Reshape}(\mathcal{F}_{i-1} \to \mathbb{R}^{\frac{TH_{i-1}W_{i-1}}{P_i^2} \times P_i^2C_{i-1}})),
\end{aligned}
\end{equation}
where $\ell_{(\cdot)}$ denotes the linear embedding layer. The feature embedding $\mathcal{F}_i\in \mathbb{R}^{(T \times H_i\times W_i) \times C_i}$ then requires processing through a multi-layer encoder. We have constructed a Transformer encoder consisting of $L$ layers, which include both attention and feedforward layers. Following \cite{pvt}, Fig.\ref{fig:ram} illustrates the Dimensional Collapse Module (DCM) that replaces the traditional Multi-Head Attention (MHA) layer in the encoder, thereby reducing computational costs. For a specific feature embedding $\mathcal{F}_i \in \mathbb{R}^{T \times H_i \times W_i \times C_i}$, we first project it into the query space $Q_i \in \mathbb{R}^{T \times H_i \times W_i \times C_i}$. For the key and value spaces, we apply a reduction rate denoted as $\varphi_i$, representing their reduction degree at stage $i$, as illustrated below:
\begin{equation}
K_i=\text{Reshape}(\mathcal{F}_i,\varphi_i)W_K\in\mathbb{R}^{\frac{TH_iW_i}{\varphi_i^2}\times C_i},
\end{equation}
\begin{equation}
V_i=\text{Reshape}(\mathcal{F}_i,\varphi_i)W_V\in\mathbb{R}^{\frac{TH_iW_i}{\varphi_i^2}\times C_i},
\end{equation}
where $\text{Reshape}(\mathcal{F}_i, \varphi_i)$ reshapes the input $\mathcal{F}_i$ into a sequence of size $\frac{TH_iW_i}{\varphi_i^2}\times(\varphi_i^2C_i)$, and $W_{(\cdot)}\in\mathbb{R}^{(\varphi_i^2C_i)\times C_i}$ is a linear projection operation that reduces the input sequence's dimensionality to $C_i$. Finally, the multi-scale features from the hierarchical Transformer are effectively integrated into the RAM to achieve feature fusion and refinement.
\subsection{2D-to-3D Back-Projection}
\label{sec:backprojection}
During the retrieval stage, VQL-2D outputs a spatio-temporal trajectory from which we extract the target's center coordinates $(u_i, v_i)$ in frame $i$ by converting the predicted segmentation mask into a bounding box. To obtain robust depth estimates in the presence of partial occlusions and to suppress noise from unreliable depth regions, we employ a confidence-weighted depth aggregation:
\begin{equation}
    \bar{\mathcal{D}}_i = \frac{\sum_{(x,y) \in \mathcal{SM}_i} \mathcal{D}_i(x,y) \cdot \text{pr}_{(x,y)}^i}{\sum_{(x,y) \in \mathcal{SM}_i} \text{pr}_{(x,y)}^i},
    \label{eq:mask_weighted_depth}
\end{equation}
where $\mathcal{SM}_i$ denotes the predicted segmentation mask, $\mathcal{D}_i(x,y)$ is the per-pixel depth prediction, and $\text{pr}_{(x,y)}^i$ is the per-pixel confidence score. By prioritizing high-confidence regions, this formulation naturally filters outliers and reduces the influence of uncertain depth estimates.

We combine the confidence-weighted depth with camera intrinsics $\mathcal{K}$ and the world-to-camera pose $\mathcal{T}_i$ to back-project the 2D point into 3D world coordinates:
\begin{equation}
    \begin{bmatrix} x_i \\ y_i \\ z_i \\ 1 \end{bmatrix} = \mathcal{T}_i \begin{bmatrix} \bar{\mathcal{D}}_i \mathcal{K}^{-1}[u_i, v_i, 1]^T \\ 1 \end{bmatrix}.
    \label{eq:backproj_3d}
\end{equation}
For depth estimation, we adopt Depth Anything \cite{depth}, which produces relative (up-to-scale) depth. Since the target output is a relative displacement offset and COLMAP-based SfM \cite{sfm} imposes a consistent global scale across frames, absolute metric depth is neither necessary nor beneficial—it would only introduce additional computational cost without improving relative localization accuracy. Critically, our work maintains a unidirectional information flow: segmentation confidence from VQL-2D modulates the geometric reliability of 3D operations, but 3D estimates do not backpropagate to modify 2D predictions. This hierarchical design achieves two objectives: (i) eliminating circular dependencies that can destabilize joint optimization, (ii) preserving a transparent input-output contract for each branch.

\subsection{Geometric-Semantic Joint Confidence-Based Multi-View Aggregation}
\label{sec:gsjc}
To enhance the robustness of 3D displacement estimation under multi-view fusion, we introduce a confidence-weighted aggregation scheme that integrates both semantic and geometric cues. The Geometric-Semantic Joint Confidence (GSJC) framework evaluates each candidate frame via four complementary factors: segmentation confidence, depth consistency, reprojection consistency, and triangulation baseline quality.

\noindent\textbf{Semantic confidence score.} Given the segmentation mask $\mathcal{SM}_i$, we compute three probability statistics that collectively characterize mask quality:
\begin{equation}
    \mathcal{P}_{\text{ave}} = \frac{1}{|\mathcal{SM}_i|}\sum_{(x,y)\in \mathcal{SM}_i}\text{pr}_{(x,y)}^i, \quad \mathcal{P}_{\text{thr}} = \frac{1}{n}\sum_{\substack{(x,y)\in \mathcal{SM}_i \\ \text{pr}_{(x,y)}^i > \wp}} \text{pr}_{(x,y)}^i, \quad \mathcal{P}_{\text{max}} = \max_{(x,y)\in \mathcal{SM}_i}\text{pr}_{(x,y)}^i,
\end{equation}
where $n$ is the count of pixels exceeding threshold $\wp=0.5$, and the three terms measure mask coverage, high-confidence concentration, and peak certainty respectively. We combine these into a unified semantic score:
\begin{equation}
    \vartheta_i^{\text{sem}} = \frac{1}{3}(\mathcal{P}_{\text{ave}} + \mathcal{P}_{\text{thr}} + \mathcal{P}_{\text{max}}),
\end{equation}
assigning equal contribution to each component.

\noindent\textbf{Depth consistency.} Reliable depth predictions should exhibit low variance within the target region:
\begin{equation}
    \sigma_{d,i}^2 = \frac{1}{|\mathcal{SM}_i|}\sum_{(x,y)\in \mathcal{SM}_i}\left(\mathcal{D}_i(x,y) - \bar{\mathcal{D}}_i\right)^2.
\end{equation}
The depth consistency weight is defined as:$\varpi_i^{\text{depth}} = \exp\left(-\frac{\sigma_{d,i}^2}{\sigma_0^2}\right)$, where $\sigma_0$ is the dataset median depth variance computed over validation frames. Frames with low within-mask variance receive weights near 1, while high-variance regions are attenuated accordingly.

\noindent\textbf{Reprojection consistency.} We verify geometric coherence by triangulating $N_C$ candidate frames. Back-projecting each using Eq.~(\ref{eq:backproj_3d}) yields 3D points $\{x_j, y_j, z_j\}_{j=1}^{N_C}$. We compute their mean:
\begin{equation}
    \bar{\mathbf{p}} = \frac{1}{N_C}\sum_{j=1}^{N_C}[x_j, y_j, z_j]^T,
\end{equation}
then reproject this consensus point to frame $i$ by first transforming to camera coordinates and then projecting:
\begin{equation}
    \bar{\mathbf{p}}_c = \mathcal{T}_i \begin{bmatrix} \bar{\mathbf{p}} \\ 1 \end{bmatrix}, \quad [\hat{u}_i, \hat{v}_i]^T = \mathcal{K} \frac{\bar{\mathbf{p}}_c^{(1:2)}}{\bar{\mathbf{p}}_c^{(3)}},
\end{equation}
where $\bar{\mathbf{p}}_c^{(1:2)}$ and $\bar{\mathbf{p}}_c^{(3)}$ denote the $xy$ and $z$ components of the camera-frame point. The reprojection weight is:
\begin{equation}
    \varpi_i^{\text{reproj}} = \exp\left(-\frac{\parallel [u_i,v_i]^T - [\hat{u}_i,\hat{v}_i]^T \parallel_2}{\epsilon_r}\right),
\end{equation}
where the tolerance is scaled by object size: $\epsilon_r = \max(10, 0.02 \times \text{bbox\_diag})$. This adaptive tolerance (typically 10--25 pixels for egocentric objects) ensures scale-dependent consistency checking. Frames whose back-projected consensus aligns well with the original 2D observation receive higher weights.

\noindent\textbf{Triangulation Baseline.} Triangulation accuracy depends fundamentally on viewing baseline geometry. We define the optical ray direction in world coordinates by first computing the ray in camera coordinates and then rotating it back to world frame. Let $\mathbf{r}_c = \mathcal{K}^{-1}[u_i, v_i, 1]^T$ denote the unnormalized ray direction in camera space. The ray direction in world coordinates is:
\begin{equation}
    \mathbf{r}_i = \frac{R_i^T \mathbf{r}_c}{\parallel R_i^T \mathbf{r}_c \parallel_2},
\end{equation}
where $R_i$ is the rotation component of $\mathcal{T}_i$. The mean ray across all candidate frames is: $\bar{\mathbf{r}} = \frac{1}{N_C}\sum_{j=1}^{N_C}\mathbf{r}_j$. The baseline quality factor is: $\varpi_i^{\text{tri}} = \sin(\theta_i) = \sqrt{1 - (\mathbf{r}_i \cdot \bar{\mathbf{r}})^2}$, where $\theta_i$ is the angle between ray $\mathbf{r}_i$ and the mean direction. To prevent numerical degeneracy when rays are nearly parallel, we clamp the result:
$\varpi_i^{\text{tri}} = \max(\sin(\theta_i), 0.01)$. Larger baseline angles enhance triangulation stability, as orthogonal rays are less sensitive to depth noise.

\noindent\textbf{Multi-View Aggregation.} We combine all confidence factors multiplicatively to obtain a composite weight:
\begin{equation}
    \tilde{\vartheta}_i = \max\left(\vartheta_i^{\text{sem}} \cdot \varpi_i^{\text{depth}} \cdot \varpi_i^{\text{reproj}} \cdot \varpi_i^{\text{tri}},\; \epsilon_w\right),
    \label{eq:gsjc_weight}
\end{equation}
where $\epsilon_w = 10^{-6}$ prevents numerical underflow. This multiplicative aggregation ensures that any significantly unreliable factor suppresses that frame's contribution, while frames robust across all dimensions receive full weight. The final 3D position is computed as a confidence-weighted mean:
\begin{equation}
    [\hat{x}, \hat{y}, \hat{z}]^T = \frac{\sum_{j=1}^{N_C} \tilde{\vartheta}_j \cdot [x_j, y_j, z_j]^T}{\sum_{j=1}^{N_C} \tilde{\vartheta}_j}.
    \label{eq:aggregation}
\end{equation}
When all frames are severely unreliable (all weights clamp to $\epsilon_w$), the denominator approaches $N_C \epsilon_w$ and the aggregation gracefully defaults to an unweighted average, ensuring numerical stability. Finally, we transform the aggregated world position into the query frame's coordinate system via the inverse query pose:
\begin{equation}
    \boldsymbol{\delta} = \mathcal{T}_{\text{query}}^{-1}[\hat{x}, \hat{y}, \hat{z}, 1]^T,
\end{equation}
yielding the VQL-3D prediction—the target's 3D offset in the query frame's local coordinate system.

\section{Experiments}
\label{sec:experiments}
\subsection{Implementation details}
\noindent\textbf{Experimental setup}. 
Video clips are uniformly preprocessed to a resolution of $448 \times 448$ via cropping and zero-padding. We employ a PVT-L\cite{pvt} as the backbone, fine-tuning the remaining network on Ego4D-VQ\cite{ego4d} and VISOR\cite{visor}\footnote{See \ref{sec:data_ablation} for an ablation study isolating architectural contributions from the VISOR}. Depth estimation using the frozen \emph{Depth Anything-Large}\cite{depth} model. The model is optimized with AdamW~\cite{adamw}, running for 20,000 iterations with a peak learning rate of $1.5\times10^{-3}$, a weight decay of 0.02, and a step learning rate scheduler that applies three successive decay factors of $\times0.1$ at iterations 2k, 4k, and 11k, preceded by a linear warm-up over the first 250 iterations. All models are trained on GTX 4090 GPUs. Following \cite{ego4d}, we augment the training data by replacing the visual crop with samples from the response track. To simulate viewpoint variations, the visual crop is randomly rotated between $-135^\circ$ and $135^\circ$. Additionally, we apply random brightness enhancements to both the video frames and the visual crop to account for varying lighting conditions. For VQL-3D, we select sharp frames using a Laplacian operator before processing them with COLMAP\cite{colmap} for camera pose estimation. We adopt the \texttt{SIMPLE\_RADIAL\_FISHEYE} camera model to address fisheye distortion in egocentric views. Sparse reconstruction is performed via \texttt{Sequential} matching with a window size of 15, followed by \texttt{Poisson} surface reconstruction using the \texttt{vocab\_tree\_flickr100K\_words1M} vocabulary tree. Finally, a post-processing step aligns the estimated COLMAP poses with the Matterport world coordinate system. Using COLMAP's \texttt{model\_aligner} function, we estimate a Sim3 transformation between the two spaces based on at least five rendered images with known camera poses from the Matterport scan. This alignment ensures that our results are evaluated in the exact coordinate system where annotators labeled the ground truth bounding boxes.

\noindent{\textbf{Training stage}}\footnote{Details regarding the training functions are provided in \ref{app:additionalablations}.}. We train the model with a composite objective $\mathcal{L}_{total}$ that combines a classification term $\mathcal{L}_{BCE}$, two regression terms($\mathcal{L}_{smooth}$ and $\mathcal{L}_{IoU}^{s2b}$), a DCF loss $\mathcal{L}_f$, and a segmentation branch loss$\mathcal{L}_{seg}$:
\begin{equation}
    \mathcal{L}_{total}=\eta_1\cdot\mathcal{L}_{BCE}+\eta_2\cdot(\mathcal{L}_{smooth}+\mathcal{L}_{IoU}^{s2b})+\eta_3\cdot\mathcal{L}_f+\eta_4\cdot\mathcal{L}_{seg},
\end{equation}
with $\eta_1=\eta_3=0.1$, $\eta_2=0.4$, $\eta_4=0.2$ set empirically.

The classification loss $\mathcal{L}_{BCE}$ is the binary cross-entropy between the predicted probability $p^{pr}_i$ and the ground-truth label $p^{gt}_i\in\{0,1\}$:
\begin{equation}
    \mathcal{L}_{BCE}=-\frac{1}{N_B}\sum_{i=1}^{N_B}[p^{gt}_i\log(p^{pr}_i)+(1-p^{gt}_i)\log(1-p^{pr}_i)],
\end{equation}
where $N_B$ is the number of samples. For bounding-box regression, we use the Smooth L1 loss
\begin{equation}
    \begin{aligned}
    &\mathcal{L}_{smooth}=\sum_{j\in\{x,y,w,h\}}\mathrm{smooth}_{L_1}(b^{pr}_j-b^{gt}_j),\\
    &\mathrm{smooth}_{L_1}(x)=\begin{cases}0.5x^2,&\mathrm{if} |x|<1\\|x|-0.5,&\mathrm{otherwise}\end{cases}
    \end{aligned}.
\end{equation}
Here, $b^{pr}_j$ and $b^{gt}_j$ are the predicted and ground-truth values of the bounding-box center coordinates and dimensions, respectively. The VQL-2D branch also provides a segmentation mask that we convert into a fitted bounding box. We first binarize the predicted mask with a threshold of 0.5. We then fit an ellipse by least squares to obtain an initial box, parameterized by the ellipse center and its major and minor axes. We refine the box by maximizing the intersection over union between the binary mask and the fitted box, using pixel counts inside and outside the box:
\begin{equation}
\mathcal{L}_{IoU}^{s2b}=1-\frac{\mathcal{N}_{\mathrm{IS}}^{+}}{\alpha_{IoU} \mathcal{N}_{\mathrm{IS}}^{-}+\mathcal{N}_{\mathrm{IS}}^{+}+\mathcal{N}_{\mathrm{OS}}^{+}},
\end{equation}
where $\mathcal{N}_{\mathrm{IS}}^+$ and $\mathcal{N}_{\mathrm{OS}}^+$ are the numbers of pixels of objects inside and outside the bounding box, respectively, and $\mathcal{N}_{\mathrm{IS}}^-$ is the number of background pixels inside the bounding box. As Ego4D-VQ lacks pixel-level mask annotations while VISOR provides dense segmentation labels for EPIC-KITCHENS egocentric videos, we adopt a mixed supervision strategy to train the segmentation branch. For VISOR samples, which account for 30\% of training iterations, we apply the complete segmentation loss $L_{\mathrm{seg}}^{\mathrm{VISOR}}$ using pixel-wise binary cross-entropy between predicted masks and ground-truth labels. For Ego4D-VQ samples, representing 70\% of iterations, we generate pseudo-masks by executing SAM with ground-truth bounding boxes as prompts. These pseudo-masks serve as weak supervision targets with a down-weighted segmentation loss coefficient $\lambda_{\mathrm{pdo}} = 0.3 \cdot \eta^4$. Additionally, $L_{\mathrm{IoU}}^{s2b}$ and $L_{\mathrm{smooth}}$ provide indirect bounding-box-level supervision to evaluate mask quality. Formally, the segmentation loss is defined as:
\begin{equation}
L_{\mathrm{seg}} = \mathbb{1}[\text{VISOR}] \cdot L_{\mathrm{BCE\_mask}} + \mathbb{1}[\text{Ego4D}] \cdot \lambda_{\mathrm{pdo}} \cdot L_{\mathrm{BCE\_pdo}},
\end{equation}
where the indicator function $\mathbb{1}[\cdot]$ switches between supervision modes based on the data source of each mini-batch sample. This design enables the model to leverage the rich mask supervision in VISOR while maintaining temporal and semantic consistency on Ego4D-VQ frames despite the absence of pixel-level ground truth. The loss $\mathcal{L}_f$ is the DCF loss defined in Eq.(\ref{eq:dcf_loss}).
\begin{table}[t]
\centering
\caption{\textbf{Comparison results on Ego4D-VQ2D.}}
\label{tab:vq2d}
\resizebox{\textwidth}{!}{%
\begin{tabular}{l cccc cccc}
\toprule
\multirow{2.5}{*}{Method} & \multicolumn{4}{c}{VQ2D Test Server} & \multicolumn{4}{c}{VQ2D Validation Set} \\ 
\cmidrule(lr){2-5} \cmidrule(lr){6-9}
& tAP$_{25}\uparrow$ & stAP$_{25}\uparrow$ & Rec.$\uparrow$ & Succ.$\uparrow$ & tAP$_{25}\uparrow$ & stAP$_{25}\uparrow$ & Rec.$\uparrow$ & Succ.$\uparrow$ \\ 
\midrule
Ego4D Baseline \cite{ego4d} & 0.20 & 0.13 & 32.20 & 39.80 & 0.22 & 0.15 & 32.92 & 43.24 \\
NFM \cite{vq2d} & 0.24 & 0.17 & 35.29 & 43.07 & 0.26 & 0.19 & 37.88 & 47.90 \\
CocoFormer \cite{wallet} & 0.25 & 0.18 & 42.34 & 48.37 & 0.26 & 0.19 & 37.67 & 47.68 \\
VQLoc \cite{vqloc} & 0.32 & 0.24 & 45.10 & 55.88 & 0.31 & 0.22 & 47.05 & 55.89 \\
HERO-VQL \cite{herovql} & 0.38 & 0.28 & 45.32 & 60.73 & 0.38 & 0.28 & 44.90 & 61.10 \\
PRVQL \cite{prvql} & 0.37 & 0.28 & 45.70 & 59.43 & 0.37 & 0.28 & 45.70 & 59.43 \\
RELOCATE \cite{relocate} & 0.43 & 0.35 & 50.55 & 60.14 & 0.41 & 0.33 & 50.50 & 58.04 \\
\midrule
\rowcolor{gray!15} 
EgoHieraLoc (Ours) & \textbf{0.44} & \textbf{0.37} & \textbf{51.28} & \textbf{61.33} & \textbf{0.43} & \textbf{0.35} & \textbf{51.52} & \textbf{62.30} \\ 
\bottomrule
\end{tabular}%
}
\end{table}
\begin{figure*}[h!]
\centering
\includegraphics[width=0.95\textwidth]{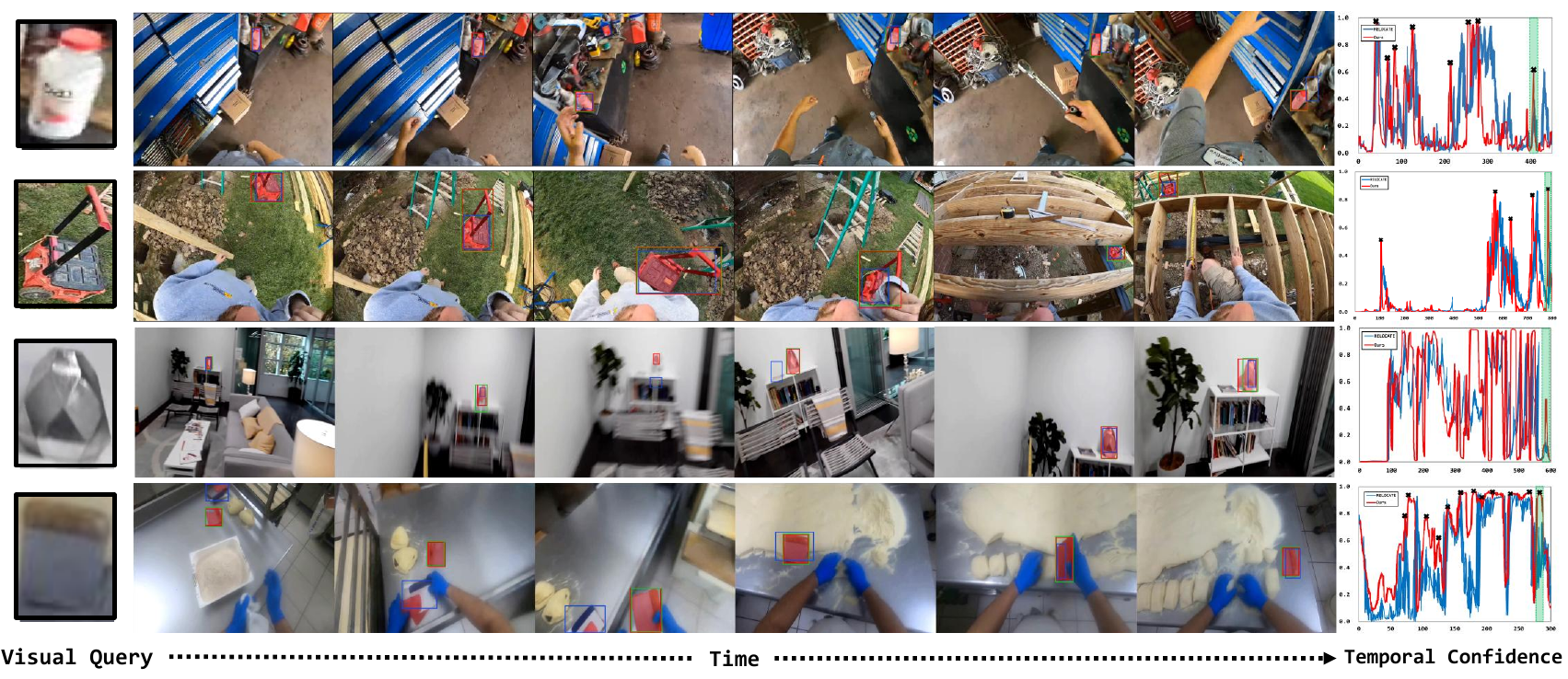}
\caption{\textbf{Qualitative visualization results on Ego4D-VQ2D.}. Each row shows a query, its corresponding response, and the temporal confidence curve of our work. \textcolor{green}{Green}, \textcolor{red}{red}, and \textcolor{blue}{blue} boxes represent the ground truth, EgoHieraLoc, and RELOCATE, respectively. The green shaded region denotes the GT interval.}
\label{fig:vis2d}
\end{figure*}
\begin{table}[t]
\centering
\caption{\textbf{Comparison results on the Ego4D-VQ3D benchmark.}}
\label{tab:vq3d}
\resizebox{\textwidth}{!}{%
\begin{tabular}{l ccccc ccccc}
\toprule
\multirow{2.5}{*}{Method} & \multicolumn{5}{c}{VQ3D Test Server (Leaderboard)} & \multicolumn{5}{c}{VQ3D Validation Set} \\ 
\cmidrule(lr){2-6} \cmidrule(lr){7-11}
& Succ.$\uparrow$ & Succ*$\uparrow$ & L2$\downarrow$ & Angle$\downarrow$ & QwP$\uparrow$ & Succ.$\uparrow$ & Succ*$\uparrow$ & L2$\downarrow$ & Angle$\downarrow$ & QwP$\uparrow$ \\ 
\midrule
Ego4D \cite{ego4d} & 7.95 & 48.61 & 4.64 & 1.31 & 0.16 & -- & -- & -- & -- & -- \\
Ego4D Improved \cite{ego4d} & 8.71 & 51.47 & 4.93 & 1.23 & 15.15 & 1.22 & 30.77 & 1.98 & 0.60 & 1.83 \\
Eivul \cite{eivul} & 25.76 & 38.74 & 8.97 & 1.21 & 66.29 & 73.78 & 91.45 & 1.35 & 0.82 & 80.49 \\
CocoFormer \cite{wallet} & 9.09 & 50.60 & 4.23 & 1.23 & 16.29 & -- & -- & -- & -- & -- \\
EgoCOL \cite{egocol} & 62.88 & 85.27 & 2.37 & \textbf{0.53} & 74.62 & 59.15 & 93.39 & 2.31 & 0.58 & 63.42 \\
EgoLoc \cite{egoloc} & 87.12 & 96.14 & 1.86 & 0.92 & 90.53 & 80.49 & 98.14 & 1.45 & 0.61 & 82.32 \\
\midrule
\rowcolor{gray!15} 
EgoHieraLoc (Ours) & \textbf{87.50} & \textbf{96.17} & \textbf{1.80} & 0.87 & \textbf{90.91} & \textbf{82.25} & \textbf{98.22} & \textbf{1.30} & \textbf{0.58} & \textbf{84.48} \\ 
\bottomrule
\end{tabular}%
}
\end{table}
\begin{figure*}[h]
\centering
\includegraphics[width=0.95\textwidth]{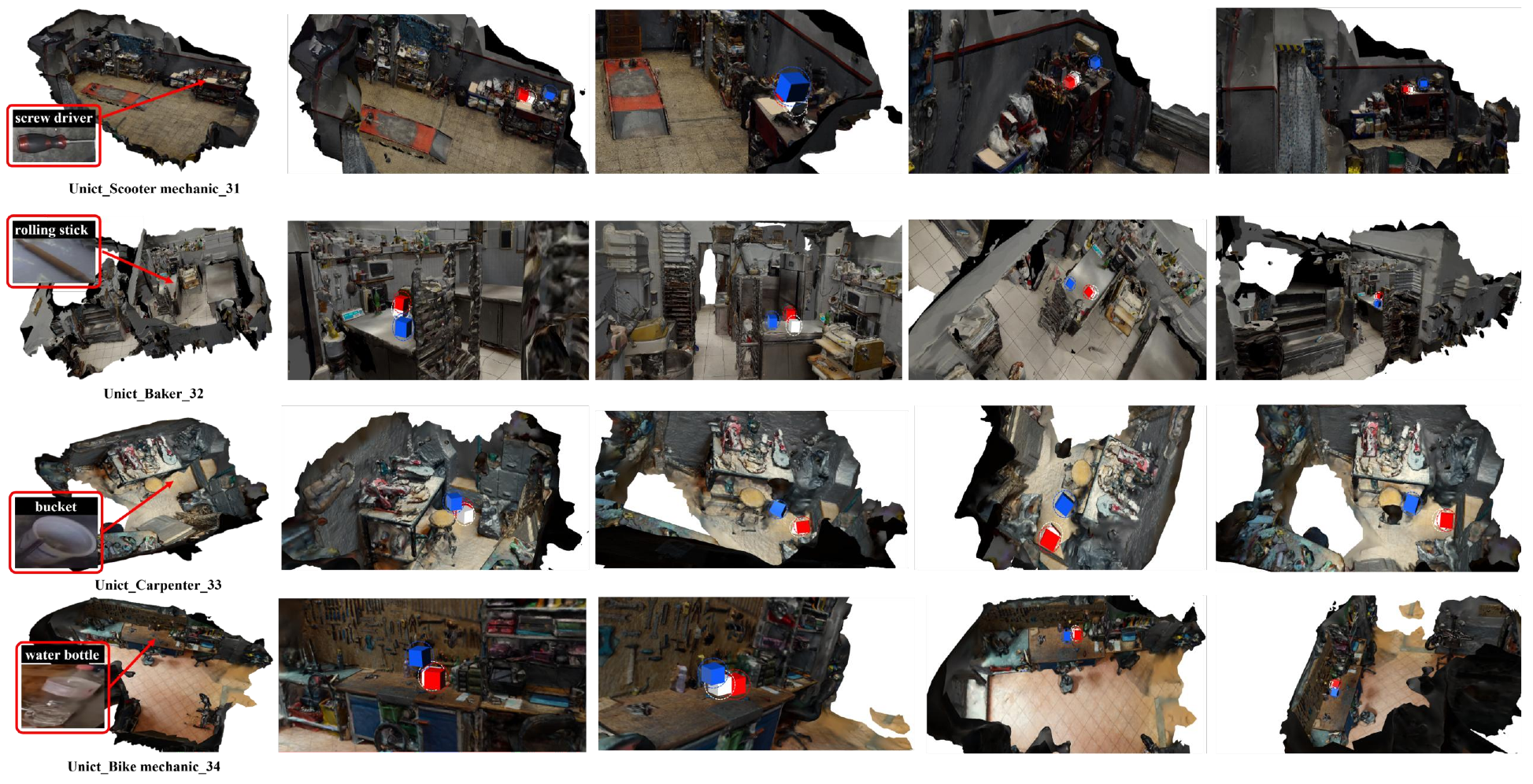}
\caption{\textbf{Visualization of VQ3D results across four reconstructed \textit{Matterport Scan} scenes.} Ground-Truth (white), EgoLoc (blue), and our predictions (red) are projected into 3D scenes reconstructed via COLMAP, with the reconstructed scan coordinate system and visual queries displayed on the left of each row. Four localization results are selected from non-fixed viewpoints for each scene to facilitate inspection. Since 3D scale and orientation are not predicted, all cubes utilize ground truth dimensions and rotations while centered at the predicted 3D coordinates.}
\label{fig:vq3dvisualize}
\end{figure*}
\begin{figure*}[!h]
\centering
\includegraphics[width=0.95\textwidth]{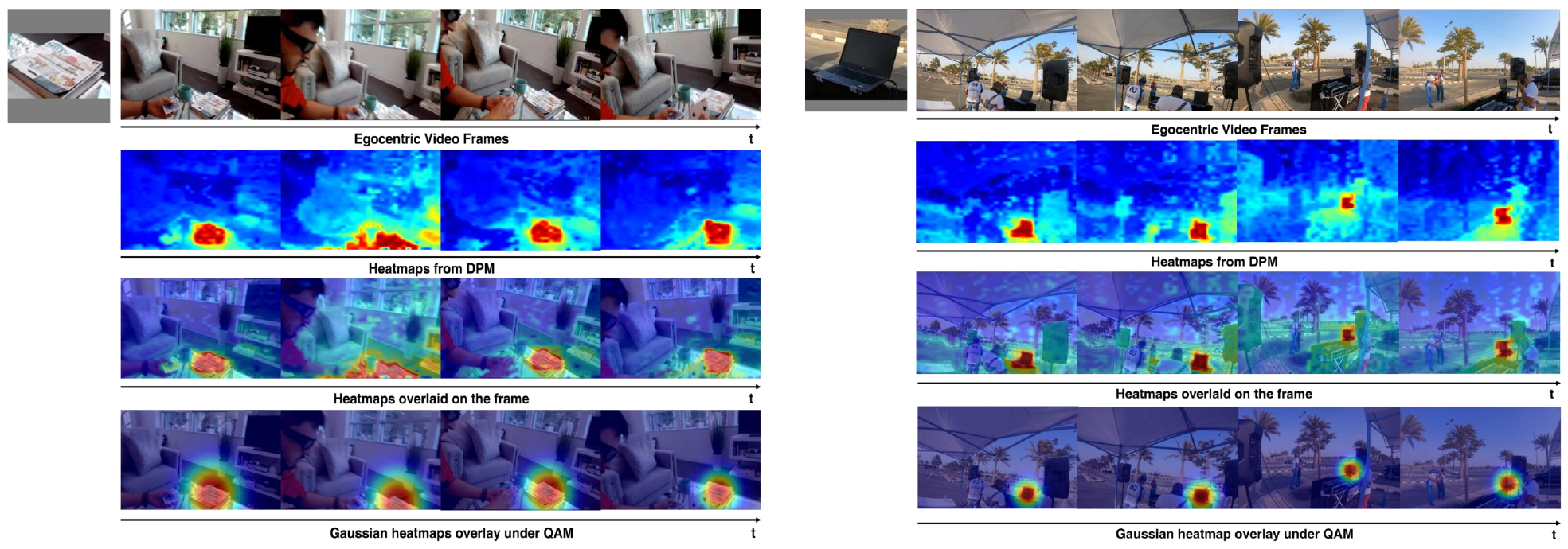}
\caption{\textbf{Qualitative analysis of heatmaps for key components.} Each example illustrates four levels of intermediate representations across temporally consecutive egocentric video frames. Row 1: egocentric video frames. Row 2: response heatmaps generated by DPM, highlighting candidate spatial regions of interest. Row 3: DPM heatmaps overlaid on corresponding frames, demonstrating spatial alignment between predicted activation and visual content. Row 4: Gaussian-smoothed heatmaps produced under QAM guidance, showing refined, semantically concentrated activation around the target object.}
\label{fig:ablatevis}
\end{figure*}

\noindent{\textbf{Inference stage}}\footnote{Inference efficiency and parameter scale are provided in \ref{app:flopsparams}.}. The video is processed in clips, and the predictions are concatenated. Given that bbox generation primarily relies on the segmentation results, we adopt the segmentation mask confidence score after aggregation, $\vartheta$, as the corresponding temporal score. To smooth the temporal scores, a median filter with a window size of $5$ is employed. Subsequently, peak detection is performed to identify the maximum peak in the matching scores. Finally, a threshold of 0.8 times the maximum peak is selected to filter the score sequence, and the last time interval in the score sequence that exceeds the threshold is retained as the temporal result for VQL-2D. For VQL-3D, we select all response tracks within this retained 2D temporal interval as our candidate frames. The center point of the bbox in each candidate frame is projected into 3D space by leveraging the corresponding frame's camera pose and depth. 3D positions are then aggregated using GSJC weighting scheme to infer the final 3D displacement of the query.
\subsection{Comparison with State-of-the-art}
We evaluate EgoHieraLoc on Ego4D-VQ\cite{ego4d}, the only publicly available benchmark for VQL. Our method is further evaluated on large-scale tracking benchmarks with minor adaptations.\footnote{See \ref{app:vql2vot} for further details on the VQL-to-VOT transfer.}

\noindent\textbf{VQL-2D (Tab.\ref{tab:vq2d}):} EgoHieraLoc achieves state-of-the-art performance across all evaluated metrics on both the VQ2D validation and test sets. Compared to RELOCATE\cite{relocate}, the proposed method consistently improves localization accuracy. On the test set, EgoHieraLoc increases tAP$_{25}$ from 0.43 to 0.44 and stAP$_{25}$ from 0.35 to 0.37, while reaching peak recall (51.28\%) and success (61.33\%) rates. These gains are even more pronounced under the stringent spatial-temporal metrics of the validation set, where EgoHieraLoc achieves a 4.26\% absolute improvement in success rate over RELOCATE (from 58.04\% to 62.30\%) and lifts recall to 51.52\%. The simultaneous increase in tAP$_{25}$ and stAP$_{25}$ suggests that the model not only retrieves correct temporal windows more reliably but also predicts tighter spatial bounding boxes within those segments. Crucially, EgoHieraLoc maintains stable performance across different data splits, avoiding the fluctuations often seen in previous models. For example, tAP$_{25}$ remains nearly identical on the validation set (0.43) and test server (0.44), with recall rates showing similar consistency (51.52\% vs. 51.28\%). This minimal degradation between the visible and hidden sets highlights the model's strong generalization capability.\footnote{Please refer to \ref{app:failure} for detailed analyses of failure cases.} This further confirms our superior capability in precise spatio-temporal localization (qualitative comparisons in Fig.\ref{fig:vis2d}).

\noindent\textbf{VQL-3D (Tab.\ref{tab:vq3d}):} EgoHieraLoc establishes a new state-of-the-art on both the VQ3D validation set and the test server leaderboard. Compared to EgoLoc\cite{egoloc}, it consistently improves across primary localization metrics. On the test server, EgoHieraLoc reaches a success rate of 87.50 and a QwP of 90.91. On the validation set, it increases the success rate from 80.49 to 82.25 and QwP from 82.32 to 84.48. EgoHieraLoc also reduces spatial localization errors; the L2 distance drops from 1.45 to 1.30 compared to EgoLoc, and the angle error reaches 0.58, matching the best baseline without sacrificing overall success. Consequently, EgoHieraLoc retrieves the object more reliably while predicting its position with higher geometric accuracy. This demonstrates enhanced robustness in camera-scene alignment and 3D localization accuracy (visualized in Fig.\ref{fig:vq3dvisualize}).

\begin{table}[t]
\caption{\textbf{Ablation studies of DPM and QAM on VQ2D and VQ3D}.}
\label{tab:ablationdpmqamdcf}
\resizebox{\textwidth}{!}{%
\begin{tabular}{@{}ccc|cccc|ccccc@{}}
\toprule
\multicolumn{2}{c}{\textbf{DPM}} & & \multicolumn{4}{c|}{\textbf{VQ2D Validation Set}} & \multicolumn{5}{c}{\textbf{VQ3D Validation Set}} \\ \cmidrule(r){1-2} \cmidrule(l){4-12} 
\textbf{$\mathcal{O}$} &
  \textbf{$\mathcal{Z}$} &
  \multirow{-2}{*}{\textbf{QAM}} &
  \textbf{tAP25}$\uparrow$ &
  \textbf{stAP25}$\uparrow$ &
  \textbf{Rec.\%}$\uparrow$ &
  \textbf{Succ.(\%)}$\uparrow$ &
  \textbf{Succ(\%)}$\uparrow$ &
  \textbf{Succ*(\%)}$\uparrow$ &
  \textbf{L2}$\downarrow$ &
  \textbf{Angle}$\downarrow$ &
  \textbf{QwP\%}$\uparrow$ \\ \midrule
\textbf{}       & \textbf{$\checkmark$}     & \textbf{$\checkmark$}  & 0.33       & 0.22       & 47.70      & 43.09      & 65.47    & 68.22    & 3.35    & 1.38   & 84.48   \\
\textbf{$\checkmark$}      & \textbf{}      & \textbf{$\checkmark$} & 0.30       & 0.19       & 42.27      & 41.22      & 63.20    & 66.38    & 3.72    & 1.32   & 84.48   \\
                &                &  \textbf{$\checkmark$} & 0.27       & 0.18       & 34.25      & 36.01      & 58.12    & 60.23    & 3.55    & 1.30   & 84.48   \\
\textbf{$\checkmark$} & \textbf{$\checkmark$}            &  & 0.37       & 0.26       & 48.27      & 51.33      & 76.44    & 93.95    & 2.08    & 1.17   & 84.48   \\
\rowcolor[HTML]{C0C0C0} \textbf{$\checkmark$}      & \textbf{$\checkmark$}     & \textbf{$\checkmark$}& 0.43       & 0.35       & 51.52      & 62.30      & 82.25    & 98.22    & 1.30    & 0.58   & 84.48   \\ \bottomrule
\end{tabular}%
}
\end{table}
\begin{figure}[t]
  \centering
  \begin{minipage}[t]{0.31\textwidth}
    \centering
    \includegraphics[width=\textwidth]{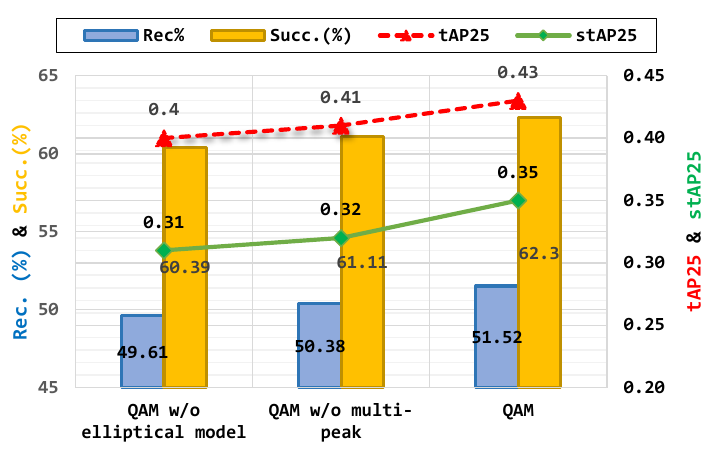}
    \centerline{(a)}
  \end{minipage}
  \hfill
  \begin{minipage}[t]{0.31\textwidth}
    \centering
    \includegraphics[width=\textwidth]{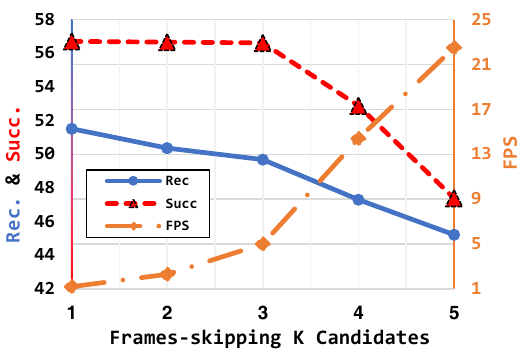}
    \centerline{(b)}
  \end{minipage}
  \begin{minipage}[t]{0.31\textwidth}
    \centering
    \includegraphics[width=\textwidth]{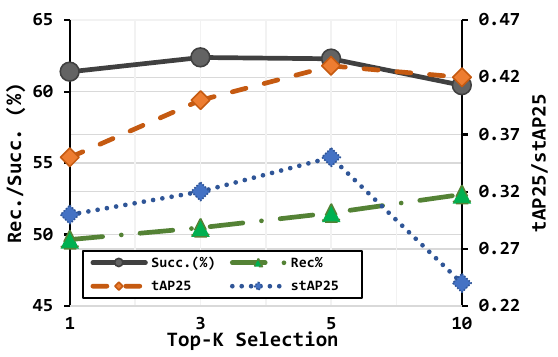}
    \centerline{(a)}
  \end{minipage}
  \caption{\textbf{Ablations on QAM components, k-frame skipping, and Top-K$_{QAM}$.} (a) Ablation studies of QAM modules on Ego4D-VQ2D. (b) Frame k-skipping ablation. Different k-values affect FPS/Success on VQ2D validation set. (c) Sensitivity analysis on the number of candidates K. On VQ2D, increasing K improves recall by recovering occluded targets, but excessive candidates (K=10) degrade localization precision (stAP).}
  \label{fig:ablations}
\end{figure}
\begin{table}[t]
\centering
\caption{\textbf{Ablation study of RAM components on Ego4D-VQ2D validation set.} 
Our default full model configuration is highlighted in \colorbox{gray!15}{light gray}.}
\label{tab:ablationstudyht}
\resizebox{\textwidth}{!}{%
\begin{tabular}{l cccc}
\toprule
\textbf{Variant / Setting} & \textbf{tAP25} $\uparrow$ & \textbf{stAP25} $\uparrow$ & \textbf{Rec. (\%)} $\uparrow$ & \textbf{Succ. (\%)} $\uparrow$ \\
\midrule
w/o DCM & 0.41 & 0.33 & 50.24 & 61.12 \\
Feature Stage: Stage 3    & 0.42 & 0.33 & 50.03 & 60.11 \\
w/o FFN & 0.38 & 0.29 & 45.74 & 59.35 \\
w/o PE & 0.43 & 0.34 & 51.04 & 61.18 \\
Backbone Architecture: ResNet-101    & 0.37 & 0.27 & 46.85 & 59.97 \\
\midrule
\rowcolor{gray!15}
\textbf{Full Model (Stage 4, w/ DCM, FFN, PE, PVT)} & \textbf{0.43} & \textbf{0.35} & \textbf{51.52} & \textbf{62.30} \\
\bottomrule
\end{tabular}%
}
\end{table}
\begin{table}[t]
\centering
\small
\setlength{\tabcolsep}{3.5pt}
\renewcommand{\arraystretch}{1.05}

\caption{\textbf{Ablation study on the multi-view aggregation function (GSJC scheme) on VQ3D validation set.} 
Semantic $\vartheta^{\text{sem}}$ and geometric $\varpi$ weights are progressively integrated. The optimal full configuration is highlighted in \colorbox{gray!15}{light gray}.}
\label{tab:aggregation}

\resizebox{\columnwidth}{!}{%
\begin{tabular}{l cccccc ccccc}
\toprule
& \multicolumn{3}{c}{\textbf{Semantic} $\vartheta^{\text{sem}}$} & \multicolumn{3}{c}{\textbf{Geometric} $\varpi$} & \multicolumn{5}{c}{\textbf{VQ3D Validation Set}} \\
\cmidrule(lr){2-4} \cmidrule(lr){5-7} \cmidrule(l){8-12}
\textbf{Variant / Setting} & $\mathcal{P}_{\text{ave}}$ & $\mathcal{P}_{\text{thr}}$ & $\mathcal{P}_{\text{max}}$ & $\varpi^{\text{depth}}$ & $\varpi^{\text{reproj}}$ & $\varpi^{\text{tri}}$ & \textbf{Succ.}$\uparrow$ & \textbf{Succ.*}$\uparrow$ & \textbf{L2}$\downarrow$ & \textbf{Ang.}$\downarrow$ & \textbf{QwP(\%)}$\uparrow$ \\
\midrule
\textit{Baseline (w/o Aggregation)} & -- & -- & -- & -- & -- & -- & 78.13 & 94.32 & 1.66 & 0.76 & 84.48 \\
\midrule
\multirow{3}{*}{\textit{Single Semantic Component}}
& \checkmark & & & & & & 78.25 & 94.40 & 1.64 & 0.72 & 84.48 \\
& & \checkmark & & & & & 78.44 & 94.93 & 1.52 & 0.67 & 84.48 \\
& & & \checkmark & & & & 79.12 & 96.13 & 1.45 & 0.64 & 84.48 \\
\midrule
\multirow{3}{*}{\textit{Semantic Combinations}}
& & \checkmark & \checkmark & & & & 80.85 & 96.78 & 1.39 & 0.61 & 84.48 \\
& \checkmark & \checkmark & & & & & 80.99 & 96.64 & 1.42 & 0.62 & 84.48 \\
& \checkmark & & \checkmark & & & & 80.95 & 96.70 & 1.40 & 0.62 & 84.48 \\
& \checkmark & \checkmark & \checkmark & & & & 81.42 & 96.93 & 1.37 & 0.60 & 84.48 \\
\midrule
\multirow{3}{*}{\textit{+ Partial Geometric Weights}}
& \checkmark & \checkmark & \checkmark & \checkmark & & & 81.63 & 97.96 & 1.32 & 0.58 & 84.48 \\
& \checkmark & \checkmark & \checkmark & & \checkmark & & 81.82 & 98.11 & 1.31 & 0.57 & 84.48 \\
& \checkmark & \checkmark & \checkmark & & & \checkmark & 81.54 & 97.84 & 1.34 & 0.56 & 84.48 \\
\midrule
\rowcolor{gray!15}
\textbf{Full GSJC Scheme} & \checkmark & \checkmark & \checkmark & \checkmark & \checkmark & \checkmark & \textbf{82.25} & \textbf{98.22} & \textbf{1.30} & \textbf{0.56} & \textbf{84.48} \\
\bottomrule
\end{tabular}%
}
\end{table}
\subsection{Ablation Study}
\label{sec:ablation}
\noindent\textbf{Impact of DPM \& QAM}. As shown in Tab.\ref{tab:ablationdpmqamdcf}, we conducted ablation studies on Ego4D-VQ2D and -VQ3D validation sets, using the following variants: (i) removal of $\mathcal{O}$ from the DPM; (ii) exclusion of the likelihood channel map $\mathcal{Z}$ from the DPM; (iii) utilization of only the QAM; and (iv) retention of only the DPM. The experimental results indicate that, compared to retaining all components, individually removing either $\mathcal{O}$ or $\mathcal{Z}$ degrades performance on both benchmarks, while their simultaneous removal leads to a more substantial performance decline, highlighting their complementarity and the overall importance of the DPM. Meanwhile, ablating the QAM module also results in varying degrees of negative impact. It can be observed that performance on 2D tasks directly influences the 3D results, and it also suggests that the DPM is slightly more critical than the QAM, although both modules play indispensable and crucial roles. Fig.\ref{fig:ablatevis} shows the heatmap visualization of the two modules and their related components overlaid on the image. It can be seen that the synergy of them makes the localization more robust and accurate.

\noindent\textbf{Impact of components of QAM}. We further analyzed the critical components of QAM. Specifically, we evaluated two QAM variants on VQ2D: (i) QAM without the elliptical model constraint, and (ii) QAM without multi-peak detection. As shown in Fig.\ref{fig:ablations}(a), removing either component leads to a performance drop. It can be observed that the elliptical model constraint serves as an effective complement to DCF, adapting better to non-rigid target deformations than bounding boxes by providing a spatial morphological prior for the object. Removing this component weakens the model's prior knowledge of the target shape, resulting in failures when the object undergoes significant deformation. Meanwhile, removing multi-peak detection introduces localization ambiguity, as the target's response map in videos may exhibit multiple salient or discontinuous peaks.

\noindent\textbf{Impact of k-frame skipping}. The k-frame skipping strategy demonstrates varying degrees of improvement in both frames per second (FPS) and localization performance. Ablation studies were conducted on the VQ2D for the k-frame skipping strategy, with the results depicted in Fig.\ref{fig:ablations}(b). When we increase the number of skipped frames k, FPS rises while performance deteriorates. We attribute this to three factors: (i) loss of temporal continuity: skipping more frames enlarges the time gap between consecutive inputs, impeding the model’s ability to capture smooth target motion and causing drift in tracking boundaries and localization points; (ii) loss of key‐frame information: key frames carry fine‐grained details for target localization; (iii) amplified appearance and contextual variation: as the frame interval grows, changes in scene illumination, viewpoint or target appearance become more pronounced, depriving the model of sufficient inter‐frame comparisons to distinguish target from background.

\noindent\textbf{Impact of Top-K$_{QAM}$}. We investigate the impact of the candidate number $K \in \{1, 3, 5, 10\}$ in the QAM module. For VQ2D, Fig.\ref{fig:ablations}(c) shows that increasing $K$ from 1 to 5 consistently improves Rec, stAP, and tAP. This validates that considering multiple peaks effectively retrieves targets under severe occlusion or motion blur. However, a larger $K=10$ introduces significant background noise, causing a sharp drop in stAP and Succ. Therefore, we select $K=5$ as the optimal setting to balance retrieval robustness and localization accuracy. 

\noindent\textbf{Impact of RAM}. 
We took several ablation studies to analyze the functional of the hierarchical Transformer backbone and evaluated some variants: (i) W/O DCM: standard Transformer lacking hierarchy without DCM. (ii) Stage[3,4]: two/four-stage hierarchical Transformers. (iii) W/O FFN: lightweight variant, removing FFN layers for reduced complexity. (iv) W/O PE: variant without position encoding. (vi) ResNet-101\cite{resnet} variant replaced the hierarchical Transformer architecture. Tab.\ref{tab:ablationstudyht} shows that replacing the DCM with a standard attention mechanism does not improve performance, likely because fine-grained feature scaling is necessary during fusion with segmentation features. While deeper hierarchical Transformers can refine representations, they offer only marginal overall gains. The impact of absolute positional encoding is similarly negligible; considering the object dynamics induced by camera motion, relative positional information may be more effective. Comparing backbone architectures, our model outperforms the ResNet-101 variant, highlighting the limitations of CNNs in capturing the long-range dependencies and global context required for cross-scale visual localization. These results reinforce the importance of Transformers in correlating global and local features, a capability that our module is designed to exploit.

\noindent\textbf{Impact on Multi-view Aggregation}\footnote{A comprehensive analysis of GSJC is provided in \ref{app:gsjc}.}. We first evaluate the impact of semantic weighting via $\vartheta^{sem}$. Results indicate that incorporating the full set of proxies ($\mathcal{P}_{ave}$, $\mathcal{P}_{thr}$, and $\mathcal{P}_{max}$) significantly enhances the Success rate from 78.13\% to 81.42\% and reduces the L2 error by 0.29. Among individual components, $\mathcal{P}_{max}$ provides the most substantial gain, demonstrating that identifying peak-response frames is crucial for filtering out noisy temporal observations. The integration of $\varpi^{depth}$, $\varpi^{reproj}$, and $\varpi^{tri}$ culminates in an optimal Success rate of 82.25\% and a Success* of 98.22\%. Notably, $\varpi^{reproj}$ contributes most to Succ* by enforcing multi-view consistency. While accuracy metrics show consistent improvement, QwP remains constant at 84.48\% as it is bounded by the preceding camera pose and depth estimation stage, confirming that our GSJC scheme primarily optimizes prediction precision through semantic-geometric cooperation.
\section{Conclusions}
\label{sec:conclusion}
We present EgoHieraLoc, a human-vision-inspired framework unifying VQL-2D and -3D to address egocentric VQL. It employs three cooperating 2D modules---Discriminative Parsing Module (DPM), Query-aware Module (QAM), and Regional Adaptation Module (RAM)---for robust, fine-grained object localization via hierarchical processing. Further, our Geometric-Semantic Joint Confidence (GSJC) weighting scheme enables reliability-driven multi-view fusion for stable 3D displacement estimation. By synergistically integrating semantic guidance and geometric reliability, EgoHieraLoc achieves state-of-the-art performance across VQL benchmarks, setting a strong new baseline.

\noindent\textbf{Acknowledgments}. This work supported by the National Natural Science Foundation of China (Grant:61672128) and from the Dalian Key Field Innovation Team Support Plan (Grant:2020RT07).
\bibliographystyle{elsarticle-num} 
\bibliography{ref}

@inproceedings{ego4d,
  title={Ego4d: Around the world in 3,000 hours of egocentric video},
  author={Grauman, Kristen and Westbury, Andrew and Byrne, Eugene and Chavis, Zachary and Furnari, Antonino and Girdhar, Rohit and Hamburger, Jackson and Jiang, Hao and Liu, Miao and Liu, Xingyu and others},
  booktitle={Proceedings of the IEEE/CVF Conference on Computer Vision and Pattern Recognition},
  pages={18995--19012},
  year={2022}
}

@article{vqloc,
  title={Single-stage visual query localization in egocentric videos},
  author={Jiang, Hanwen and Ramakrishnan, Santhosh Kumar and Grauman, Kristen},
  journal={Advances in Neural Information Processing Systems},
  volume={36},
  year={2024}
}

@inproceedings{egoloc,
  title={Egoloc: Revisiting 3d object localization from egocentric videos with visual queries},
  author={Mai, Jinjie and Hamdi, Abdullah and Giancola, Silvio and Zhao, Chen and Ghanem, Bernard},
  booktitle={Proceedings of the IEEE/CVF International Conference on Computer Vision},
  pages={45--57},
  year={2023}
}

@inproceedings{wallet,
  title={Where is my wallet? modeling object proposal sets for egocentric visual query localization},
  author={Xu, Mengmeng and Li, Yanghao and Fu, Cheng-Yang and Ghanem, Bernard and Xiang, Tao and P{\'e}rez-R{\'u}a, Juan-Manuel},
  booktitle={Proceedings of the IEEE/CVF Conference on Computer Vision and Pattern Recognition},
  pages={2593--2603},
  year={2023}
}

@article{eivul,
  title={Estimating more camera poses for ego-centric videos is essential for VQ3D},
  author={Mai, Jinjie and Zhao, Chen and Hamdi, Abdullah and Giancola, Silvio and Ghanem, Bernard},
  journal={arXiv preprint arXiv:2211.10284},
  year={2022}
}

@article{egocol,
  title={EgoCOL: Egocentric Camera pose estimation for Open-world 3D object Localization@ Ego4D challenge 2023},
  author={Forigua, Cristhian and Escobar, Maria and Pont-Tuset, Jordi and Maninis, Kevis-Kokitsi and Arbel{\'a}ez, Pablo},
  journal={arXiv preprint arXiv:2306.16606},
  year={2023}
}

@article{egotracks,
  title={Egotracks: A long-term egocentric visual object tracking dataset},
  author={Tang, Hao and Liang, Kevin J and Grauman, Kristen and Feiszli, Matt and Wang, Weiyao},
  journal={Advances in Neural Information Processing Systems},
  volume={36},
  year={2024}
}

@article{vq2d,
  title={Negative Frames Matter in Egocentric Visual Query 2D Localization},
  author={Xu, Mengmeng and Fu, Cheng-Yang and Li, Yanghao and Ghanem, Bernard and Perez-Rua, Juan-Manuel and Xiang, Tao},
  journal={arXiv preprint arXiv:2208.01949},
  year={2022}
}

@inproceedings{siamrcnn,
  title={Siam r-cnn: Visual tracking by re-detection},
  author={Voigtlaender, Paul and Luiten, Jonathon and Torr, Philip HS and Leibe, Bastian},
  booktitle={Proceedings of the IEEE/CVF conference on computer vision and pattern recognition},
  pages={6578--6588},
  year={2020}
}

@inproceedings{siamrpn++,
  title={Siamrpn++: Evolution of siamese visual tracking with very deep networks},
  author={Li, Bo and Wu, Wei and Wang, Qiang and Zhang, Fangyi and Xing, Junliang and Yan, Junjie},
  booktitle={Proceedings of the IEEE/CVF conference on computer vision and pattern recognition},
  pages={4282--4291},
  year={2019}
}

@inproceedings{siamrpn,
  title={High performance visual tracking with siamese region proposal network},
  author={Li, Bo and Yan, Junjie and Wu, Wei and Zhu, Zheng and Hu, Xiaolin},
  booktitle={Proceedings of the IEEE conference on computer vision and pattern recognition},
  pages={8971--8980},
  year={2018}
}

@inproceedings{d3s,
  title={D3s-a discriminative single shot segmentation tracker},
  author={Lukezic, Alan and Matas, Jiri and Kristan, Matej},
  booktitle={Proceedings of the IEEE/CVF conference on computer vision and pattern recognition},
  pages={7133--7142},
  year={2020}
}

@article{siammask,
  title={Siammask: A framework for fast online object tracking and segmentation},
  author={Hu, Weiming and Wang, Qiang and Zhang, Li and Bertinetto, Luca and Torr, Philip HS},
  journal={IEEE Transactions on Pattern Analysis and Machine Intelligence},
  volume={45},
  number={3},
  pages={3072--3089},
  year={2023},
  publisher={IEEE}
}

@inproceedings{isfirst,
  title={Is first person vision challenging for object tracking?},
  author={Dunnhofer, Matteo and Furnari, Antonino and Farinella, Giovanni Maria and Micheloni, Christian},
  booktitle={Proceedings of the IEEE/CVF International Conference on Computer Vision},
  pages={2698--2710},
  year={2021}
}

@article{egovot,
  title={Visual object tracking in first person vision},
  author={Dunnhofer, Matteo and Furnari, Antonino and Farinella, Giovanni Maria and Micheloni, Christian},
  journal={International Journal of Computer Vision},
  volume={131},
  number={1},
  pages={259--283},
  year={2023},
  publisher={Springer}
}

@article{visor,
  title={Epic-kitchens visor benchmark: Video segmentations and object relations},
  author={Darkhalil, Ahmad and Shan, Dandan and Zhu, Bin and Ma, Jian and Kar, Amlan and Higgins, Richard and Fidler, Sanja and Fouhey, David and Damen, Dima},
  journal={Advances in Neural Information Processing Systems},
  volume={35},
  pages={13745--13758},
  year={2022}
}

@inproceedings{siamfc,
  title={Fully-convolutional siamese networks for object tracking},
  author={Bertinetto, Luca and Valmadre, Jack and Henriques, Joao F and Vedaldi, Andrea and Torr, Philip HS},
  booktitle={Computer Vision--ECCV 2016 Workshops: Amsterdam, The Netherlands, October 8-10 and 15-16, 2016, Proceedings, Part II 14},
  pages={850--865},
  year={2016},
  organization={Springer}
}

@inproceedings{stark,
  title={Learning spatio-temporal transformer for visual tracking},
  author={Yan, Bin and Peng, Houwen and Fu, Jianlong and Wang, Dong and Lu, Huchuan},
  booktitle={Proceedings of the IEEE/CVF international conference on computer vision},
  pages={10448--10457},
  year={2021}
}

@article{ masktrack,
  author = {Linjie Yang and Yuchen Fan and Ning Xu},
  title = {Video instance segmentation},
  journal = {CoRR},
  volume = {abs/1905.04804},
  year = {2019},
  url = {https://arxiv.org/abs/1905.04804}
}

@inproceedings{feelvos,
  title={Feelvos: Fast end-to-end embedding learning for video object segmentation},
  author={Voigtlaender, Paul and Chai, Yuning and Schroff, Florian and Adam, Hartwig and Leibe, Bastian and Chen, Liang-Chieh},
  booktitle={Proceedings of the IEEE/CVF conference on computer vision and pattern recognition},
  pages={9481--9490},
  year={2019}
}

@inproceedings{vistr,
  title={End-to-end video instance segmentation with transformers},
  author={Wang, Yuqing and Xu, Zhaoliang and Wang, Xinlong and Shen, Chunhua and Cheng, Baoshan and Shen, Hao and Xia, Huaxia},
  booktitle={Proceedings of the IEEE/CVF conference on computer vision and pattern recognition},
  pages={8741--8750},
  year={2021}
}

@inproceedings{seqformer,
  title={Seqformer: Sequential transformer for video instance segmentation},
  author={Wu, Junfeng and Jiang, Yi and Bai, Song and Zhang, Wenqing and Bai, Xiang},
  booktitle={European Conference on Computer Vision},
  pages={553--569},
  year={2022},
  organization={Springer}
}

@inproceedings{sam,
  title={Segment anything},
  author={Kirillov, Alexander and Mintun, Eric and Ravi, Nikhila and Mao, Hanzi and Rolland, Chloe and Gustafson, Laura and Xiao, Tete and Whitehead, Spencer and Berg, Alexander C and Lo, Wan-Yen and others},
  booktitle={Proceedings of the IEEE/CVF International Conference on Computer Vision},
  pages={4015--4026},
  year={2023}
}

@article{sam2,
  title={Sam 2: Segment anything in images and videos},
  author={Ravi, Nikhila and Gabeur, Valentin and Hu, Yuan-Ting and Hu, Ronghang and Ryali, Chaitanya and Ma, Tengyu and Khedr, Haitham and R{\"a}dle, Roman and Rolland, Chloe and Gustafson, Laura and others},
  journal={arXiv preprint arXiv:2408.00714},
  year={2024}
}

@article{nerf,
  title={Nerf: Representing scenes as neural radiance fields for view synthesis},
  author={Mildenhall, Ben and Srinivasan, Pratul P and Tancik, Matthew and Barron, Jonathan T and Ramamoorthi, Ravi and Ng, Ren},
  journal={Communications of the ACM},
  volume={65},
  number={1},
  pages={99--106},
  year={2021},
  publisher={ACM New York, NY, USA}
}

@article{3dgs,
  title={3D Gaussian Splatting for Real-Time Radiance Field Rendering.},
  author={Kerbl, Bernhard and Kopanas, Georgios and Leimk{\"u}hler, Thomas and Drettakis, George},
  journal={ACM Trans. Graph.},
  volume={42},
  number={4},
  pages={139--1},
  year={2023}
}

@inproceedings{egoslam,
  title={Ego-slam: A robust monocular slam for egocentric videos},
  author={Patra, Suvam and Gupta, Kartikeya and Ahmad, Faran and Arora, Chetan and Banerjee, Subhashis},
  booktitle={2019 IEEE Winter Conference on Applications of Computer Vision (WACV)},
  pages={31--40},
  year={2019},
  organization={IEEE}
}

@inproceedings{neuraldiff,
  title={NeuralDiff: Segmenting 3D objects that move in egocentric videos},
  author={Tschernezki, Vadim and Larlus, Diane and Vedaldi, Andrea},
  booktitle={2021 International Conference on 3D Vision (3DV)},
  pages={910--919},
  year={2021},
  organization={IEEE}
}

@inproceedings{dust3r,
  title={Dust3r: Geometric 3d vision made easy},
  author={Wang, Shuzhe and Leroy, Vincent and Cabon, Yohann and Chidlovskii, Boris and Revaud, Jerome},
  booktitle={Proceedings of the IEEE/CVF Conference on Computer Vision and Pattern Recognition},
  pages={20697--20709},
  year={2024}
}

@article{egolifter,
  title={EgoLifter: Open-world 3D Segmentation for Egocentric Perception},
  author={Gu, Qiao and Lv, Zhaoyang and Frost, Duncan and Green, Simon and Straub, Julian and Sweeney, Chris},
  journal={arXiv preprint arXiv:2403.18118},
  year={2024}
}

@article{egogaussian,
  title={EgoGaussian: Dynamic Scene Understanding from Egocentric Video with 3D Gaussian Splatting},
  author={Zhang, Daiwei and Li, Gengyan and Li, Jiajie and Bressieux, Micka{\"e}l and Hilliges, Otmar and Pollefeys, Marc and Van Gool, Luc and Wang, Xi},
  journal={arXiv preprint arXiv:2406.19811},
  year={2024}
}

@article{sfm,
  title={The interpretation of structure from motion},
  author={Ullman, Shimon},
  journal={Proceedings of the Royal Society of London. Series B. Biological Sciences},
  volume={203},
  number={1153},
  pages={405--426},
  year={1979},
  publisher={The Royal Society London}
}

@article{egooutlook,
  title={An outlook into the future of egocentric vision},
  author={Plizzari, Chiara and Goletto, Gabriele and Furnari, Antonino and Bansal, Siddhant and Ragusa, Francesco and Farinella, Giovanni Maria and Damen, Dima and Tommasi, Tatiana},
  journal={International Journal of Computer Vision},
  pages={1--57},
  year={2024},
  publisher={Springer}
}

@inproceedings{pvt,
  title={Pyramid vision transformer: A versatile backbone for dense prediction without convolutions},
  author={Wang, Wenhai and Xie, Enze and Li, Xiang and Fan, Deng-Ping and Song, Kaitao and Liang, Ding and Lu, Tong and Luo, Ping and Shao, Ling},
  booktitle={Proceedings of the IEEE/CVF international conference on computer vision},
  pages={568--578},
  year={2021}
}

@inproceedings{atom,
  title={Atom: Accurate tracking by overlap maximization},
  author={Danelljan, Martin and Bhat, Goutam and Khan, Fahad Shahbaz and Felsberg, Michael},
  booktitle={Proceedings of the IEEE/CVF conference on computer vision and pattern recognition},
  pages={4660--4669},
  year={2019}
}

@inproceedings{depth,
  title={Depth anything: Unleashing the power of large-scale unlabeled data},
  author={Yang, Lihe and Kang, Bingyi and Huang, Zilong and Xu, Xiaogang and Feng, Jiashi and Zhao, Hengshuang},
  booktitle={Proceedings of the IEEE/CVF Conference on Computer Vision and Pattern Recognition},
  pages={10371--10381},
  year={2024}
}

@inproceedings{colmap,
  title={Structure-from-motion revisited},
  author={Schonberger, Johannes L and Frahm, Jan-Michael},
  booktitle={Proceedings of the IEEE conference on computer vision and pattern recognition},
  pages={4104--4113},
  year={2016}
}

@inproceedings{dimp,
  title={Learning discriminative model prediction for tracking},
  author={Bhat, Goutam and Danelljan, Martin and Gool, Luc Van and Timofte, Radu},
  booktitle={Proceedings of the IEEE/CVF international conference on computer vision},
  pages={6182--6191},
  year={2019}
}

@inproceedings{siamfcplus,
  title={SiamFC++: Towards robust and accurate visual tracking with target estimation guidelines},
  author={Xu, Yinda and Wang, Zeyu and Li, Zuoxin and Yuan, Ye and Yu, Gang},
  booktitle={Proceedings of the AAAI conference on artificial intelligence},
  volume={34},
  number={07},
  pages={12549--12556},
  year={2020}
}

@inproceedings{globaltrack,
  title={Globaltrack: A simple and strong baseline for long-term tracking},
  author={Huang, Lianghua and Zhao, Xin and Huang, Kaiqi},
  booktitle={Proceedings of the AAAI conference on artificial intelligence},
  volume={34},
  number={07},
  pages={11037--11044},
  year={2020}
}

@inproceedings{LTMU,
  title={High-performance long-term tracking with meta-updater},
  author={Dai, Kenan and Zhang, Yunhua and Wang, Dong and Li, Jianhua and Lu, Huchuan and Yang, Xiaoyun},
  booktitle={Proceedings of the IEEE/CVF conference on computer vision and pattern recognition},
  pages={6298--6307},
  year={2020}
}

@inproceedings{tomp,
  title={Transforming model prediction for tracking},
  author={Mayer, Christoph and Danelljan, Martin and Bhat, Goutam and Paul, Matthieu and Paudel, Danda Pani and Yu, Fisher and Van Gool, Luc},
  booktitle={Proceedings of the IEEE/CVF conference on computer vision and pattern recognition},
  pages={8731--8740},
  year={2022}
}

@inproceedings{mixformer,
  title={Mixformer: End-to-end tracking with iterative mixed attention},
  author={Cui, Yutao and Jiang, Cheng and Wang, Limin and Wu, Gangshan},
  booktitle={Proceedings of the IEEE/CVF conference on computer vision and pattern recognition},
  pages={13608--13618},
  year={2022}
}

@inproceedings{resnet,
  title={Deep residual learning for image recognition},
  author={He, Kaiming and Zhang, Xiangyu and Ren, Shaoqing and Sun, Jian},
  booktitle={Proceedings of the IEEE conference on computer vision and pattern recognition},
  pages={770--778},
  year={2016}
}

@inproceedings{dcf,
  title={Visual object tracking using adaptive correlation filters},
  author={Bolme, David S and Beveridge, J Ross and Draper, Bruce A and Lui, Yui Man},
  booktitle={2010 IEEE computer society conference on computer vision and pattern recognition},
  pages={2544--2550},
  year={2010},
  organization={IEEE}
}

@misc{herovql,
    title={HERO-VQL: Hierarchical, Egocentric and Robust Visual Query Localization},
    author={Joohyun Chang and Soyeon Hong and Hyogun Lee and Seong Jong Ha and Dongho Lee and Seong Tae Kim and Jinwoo Choi},
    year={2025},
    eprint={2509.00385},
    archivePrefix={arXiv},
    primaryClass={cs.CV}
}

@inproceedings{relocate,
  title={Relocate: A simple training-free baseline for visual query localization using region-based representations},
  author={Khosla, Savya and Schwing, Alexander and Hoiem, Derek and others},
  booktitle={Proceedings of the Computer Vision and Pattern Recognition Conference},
  pages={3697--3706},
  year={2025}
}

@article{prvql,
  title={PRVQL: Progressive Knowledge-guided Refinement for Robust Egocentric Visual Query Localization},
  author={Fan, Bing and Feng, Yunhe and Tian, Yapeng and Lin, Yuewei and Huang, Yan and Fan, Heng},
  journal={arXiv preprint arXiv:2502.07707},
  year={2025}
}

@inproceedings{adamw,
  title={Decoupled Weight Decay Regularization},
  author={Loshchilov, Ilya and Hutter, Frank},
  booktitle={International Conference on Learning Representations}
}

@article{srrt,
  title={SRRT: Exploring search region regulation for visual object tracking},
  author={Zhu, Jiawen and Chen, Xin and Zhang, Pengyu and Wang, Xinying and Wang, Dong and Zhao, Wenda and Lu, Huchuan},
  journal={IEEE Transactions on Circuits and Systems for Video Technology},
  volume={34},
  number={11},
  pages={10551--10563},
  year={2024},
  publisher={IEEE}
}

@article{lgtrack,
  title={LGTrack: Exploiting local and global properties for robust visual tracking},
  author={Liu, Chang and Zhao, Jie and Bo, Chunjuan and Li, Shengming and Wang, Dong and Lu, Huchuan},
  journal={IEEE Transactions on Circuits and Systems for Video Technology},
  volume={34},
  number={9},
  pages={8161--8171},
  year={2024},
  publisher={IEEE}
}

@article{zsl,
  title={Graph-regularized structured support vector machine for object tracking},
  author={Zhang, Shunli and Sui, Yao and Zhao, Sicong and Zhang, Li},
  journal={IEEE Transactions on Circuits and Systems for Video Technology},
  volume={27},
  number={6},
  pages={1249--1262},
  year={2015},
  publisher={IEEE}
}

@inproceedings{biovql,
  title={BiOVQL: Brain-inspired One-stage Egocentric Visual Query Localization},
  author={Cao, Yifei and Wang, Guolong and Hou, Mingliang and Yu, Jizhe and Zhang, Xianjie and Bu, Xiya and Li, Zhizhen and Liu, Yu},
  booktitle={Proceedings of the 2026 International Conference on Multimedia Retrieval},
  pages={88--98},
  year={2026}
}

@inproceedings{eagle,
  title={EAGLE: Episodic Appearance-and Geometry-aware Memory for Unified 2D-3D Visual Query Localization in Egocentric Vision},
  author={Cao, Yifei and Liu, Yu and Wang, Guolong and Liu, Zhu and Wang, Kai and Zhang, Xianjie and Yu, Jizhe and Tu, Xun},
  booktitle={Proceedings of the AAAI Conference on Artificial Intelligence},
  volume={40},
  number={4},
  pages={2634--2642},
  year={2026}
}

\clearpage
\newpage
\appendix
\section*{Supplementary Material}
\section{Statistical Significance Validation}
\label{app:statistical_validation}
To demonstrate that our performance gains are statistically robust and not an artifact of random seed selection, we conduct multiple independent runs to validate the statistical significance of our results. 
Specifically, we execute 5 independent training runs with different random seeds on the Ego4D-VQ2D validation set (Table \ref{tab:multirun_vq2d}). We also measure the multi-run variance of the RELOCATE baseline under identical experimental conditions (Table \ref{tab:relocate_variance}). 

The 95\% confidence interval (CI) for the success rate improvement ($\Delta_{\text{Succ}}$) is calculated as:
\begin{equation}
\Delta_{\text{Succ}} = (\overline{\text{Succ}}_{\text{Ours}} - \overline{\text{Succ}}_{\text{RELOCATE}}) \pm 1.96 \sqrt{\frac{\sigma_{\text{Ours}}^2 + \sigma_{\text{RELOCATE}}^2}{5}}
\end{equation}
This yields $\Delta_{\text{Succ}} = 62.44\% - 58.06\% = 4.38\% \pm 0.57\%$ [95\% CI: $3.81\%$--$4.95\%$] ($p < 0.001$ via Welch's $t$-test). Since the confidence interval is strictly above zero and well exceeds the benchmark's typical noise magnitude, the observed improvement is statistically significant. This confirms that our architecture consistently outperforming the baseline is robust across random initializations.

\begin{table}[htbp]
  \centering
  \caption{Multi-run validation results across 5 independent runs with different random seeds.}
  \label{tab:combined_multirun}
  
  \begin{minipage}[t]{0.49\textwidth}
    \centering
    \subcaption{Ego4D-VQ2D Validation Set (Ours)}
    \label{tab:multirun_vq2d}
    \resizebox{\linewidth}{!}{%
      \begin{tabular}{lccccc|c}
        \toprule
        Metric & Run 1 & Run 2 & Run 3 & Run 4 & Run 5 & Mean $\pm$ Std \\
        \midrule
        tAP25    & 0.42  & 0.44  & 0.43  & 0.43  & 0.44  & $0.432 \pm 0.008$ \\
        stAP25   & 0.34  & 0.36  & 0.35  & 0.35  & 0.36  & $0.352 \pm 0.008$ \\
        Rec. (\%) & 50.95 & 51.87 & 51.23 & 51.52 & 51.65 & $51.44 \pm 0.32$ \\
        Succ. (\%)& 61.65 & 63.12 & 62.18 & 62.30 & 62.95 & $62.44 \pm 0.62$ \\
        \bottomrule
      \end{tabular}%
    }
  \end{minipage}%
  \hfill 
  \begin{minipage}[t]{0.49\textwidth}
    \centering
    \subcaption{RELOCATE Baseline Multi-run Validation}
    \label{tab:relocate_variance}
    \resizebox{\linewidth}{!}{%
      \begin{tabular}{lccccc|c}
        \toprule
        Metric & Run 1 & Run 2 & Run 3 & Run 4 & Run 5 & Mean $\pm$ Std \\
        \midrule
        Succ. (\%) & 57.85 & 58.34 & 57.92 & 58.15 & 58.06 & $58.06 \pm 0.18$ \\
        \bottomrule
      \end{tabular}%
    }
  \end{minipage}
\end{table}
\section{VQL-to-VOT Transfer}
\label{app:vql2vot}
To the best of our knowledge, Ego4D-VQ\cite{ego4d} is currently the only benchmark for VQL. The workflow of VQL shares both similarities and differences with visual object tracking (VOT). The main distinctions lie in their initialization and output logic. The core of our method (specifically the VQL-2D branch) is a segmentation-based tracking paradigm; therefore, in addition to its use for VQL, our framework can serve as a robust tracking algorithm for evaluation on egocentric tracking benchmarks. For VQL, our method initializes a segmentation mask based on an input query image. This mask is then propagated and updated frame-by-frame throughout the video in a manner similar to Video Object Segmentation (VOS) until the end of the video to determine the final appearance of the target. When tracking (mask propagation) fails or the target temporarily disappears, instead of using a separate detector to re-detect the bounding box, we attempt to “re-awaken” or “match” the segmented target in subsequent frames using the original query image and updated templates. To adapt our method for egocentric VOT, we discard the reliance on an external static query image. It directly uses the interior of the bounding box provided in the initial frame of the VOT task as a coarse initial segmentation mask. Concurrently, the logic for determining the "last appearance" is removed, and the system is adjusted to continuously output the segmentation mask for each frame. These masks are then converted into minimum bounding rectangles to complete the tracking output.

\noindent\textbf{EgoTracks.} We evaluate on the long-term egocentric benchmark EgoTracks~\cite{egotracks} under both VCT-VS and VCT-VC settings (Table.\ref{tab:egotracks}). Our method outperforms the previous SOTA EgoSTARK\cite{stark,egotracks} in both tracks. Specifically, we achieve gains of 1.2\% AO / 3.4\% F-score on VCT-VS and 5.5\% AO / 4.5\% F-score on VCT-VC. These results demonstrate that our model maintains robust tracking in long sequences solely through query appearance priors, without explicit detection modules, while significantly enhancing local feature representation and consistency (qualitative results in Fig.~\ref{fig:egotracks}).

\begin{figure*}[!h]
\centering
\includegraphics[width=\textwidth]{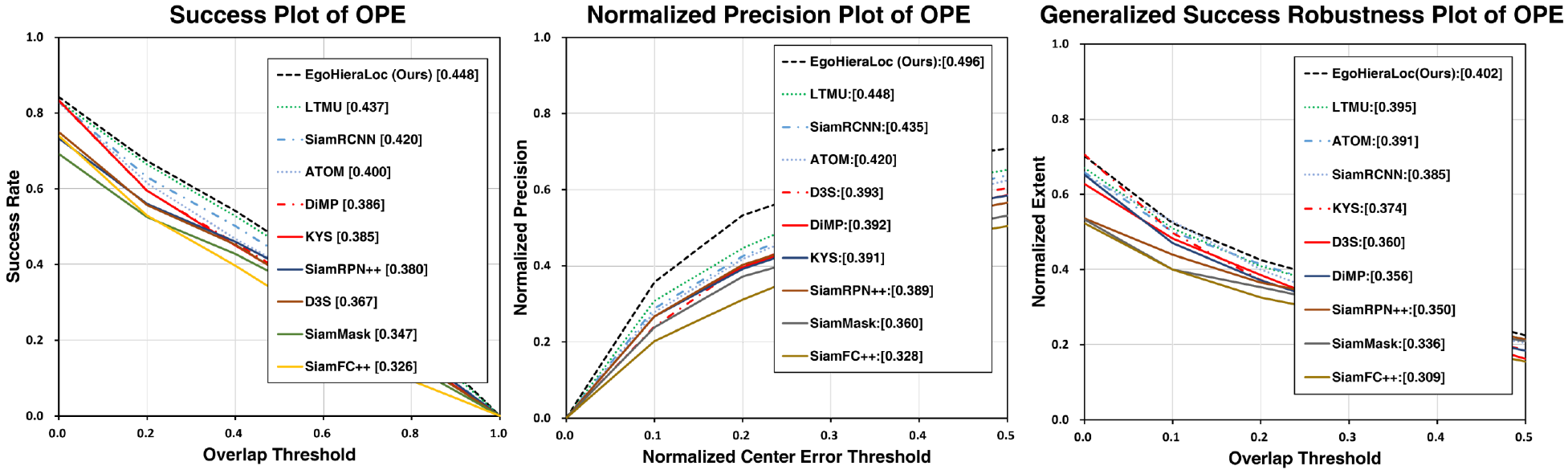}
\caption{\textbf{Comparison with the state-of-the-art trackers on TREK-150 under the OPE}. Within the brackets adjacent to the tracker names, we show the SS, NPS, and GSR values achieved by trackers among benchmarks.}
\label{fig:trekope}
\end{figure*}

\begin{figure*}[h!]
\centering
\includegraphics[width=\textwidth]{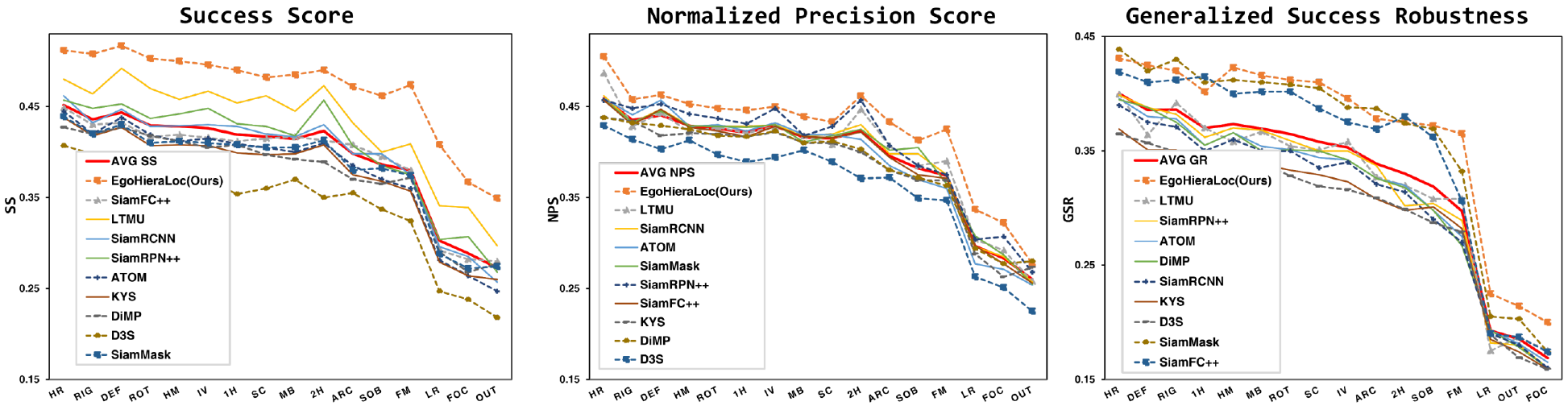}
\caption{\textbf{Comparison with the state-of-the-art trackers on the sequence attributes of TREK-150 under the MSE}.}
\label{fig:TREKMSE}
\end{figure*}

\begin{figure*}[h]
\centering
\includegraphics[width=0.9\textwidth]{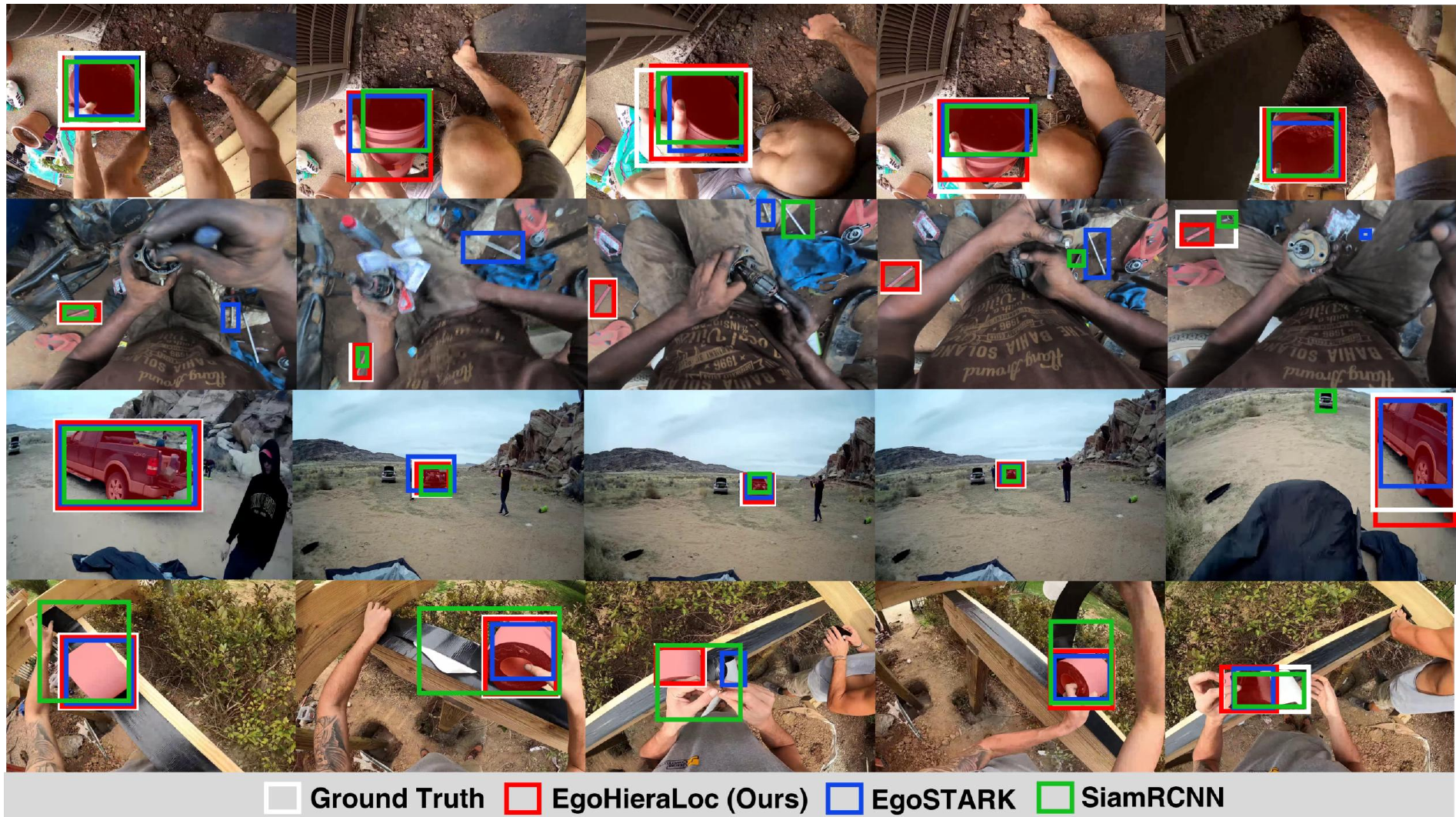}
\caption{\textbf{Visualization of comparison with SOTA trackers on EgoTracks.}}
\label{fig:egotracks}
\end{figure*}

\noindent\textbf{TREK-150.} We evaluate on the TREK-150~\cite{isfirst} benchmark using standard OPE and MSE protocols. \textbf{OPE (Fig.\ref{fig:trekope}):} EgoHieraLoc outperforms the previous best LTMU\cite{LTMU} across all metrics, achieving gains of 2.5\% (SS), 10.7\% (NPS), and 1.8\% (GSR). \textbf{MSE (Fig.~\ref{fig:TREKMSE}):} Our method demonstrates superior robustness on challenging attributes, specifically FM, LR, FOC, and OUT. This confirms EgoHieraLoc's effectiveness in handling severe viewpoint changes and occlusions characteristic of egocentric videos.

\begin{figure*}[!h]
\centering
\includegraphics[width=0.95\textwidth]{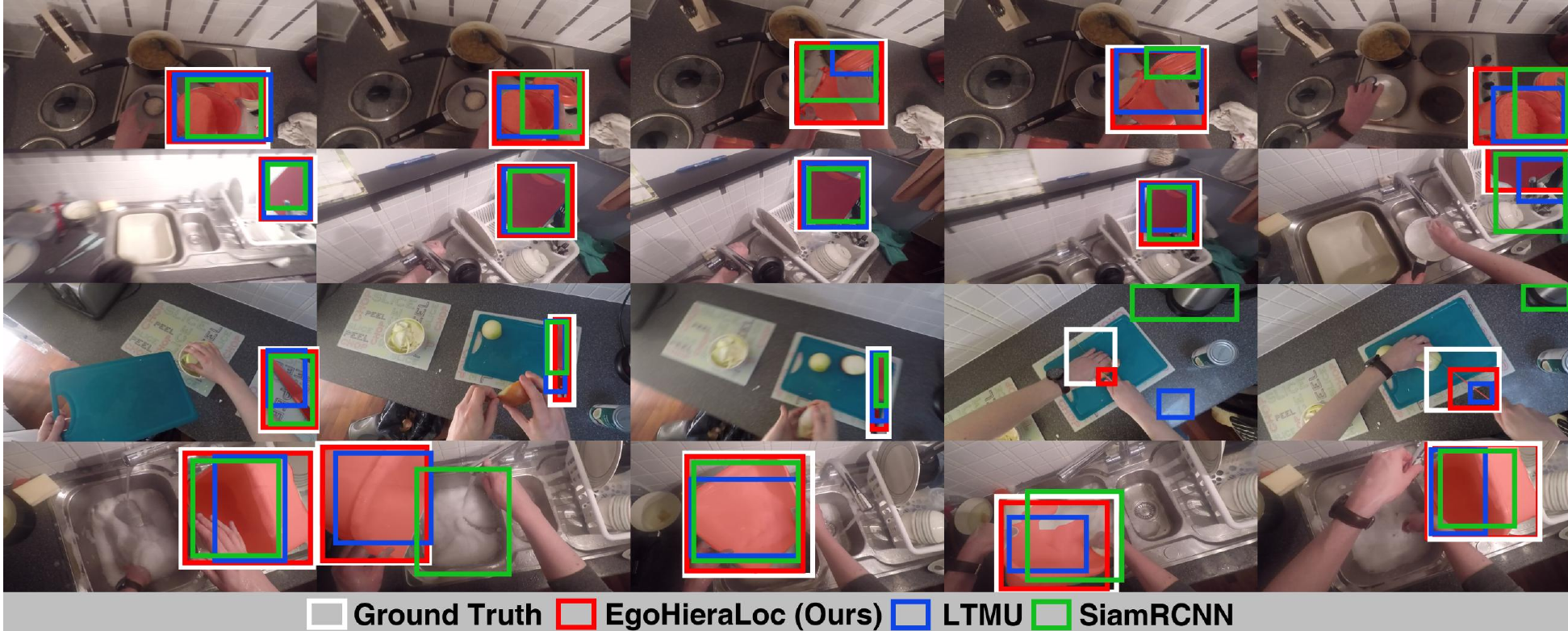}
\caption{\textbf{Qualitative comparison with state-of-the-art trackers on TREK-150.}.}
\label{fig:trek}
\end{figure*}

Fig.\ref{fig:egotracks} and Fig.~\ref{fig:trek} present qualitative visualizations of our algorithm's performance on the tracking benchmark, comparing it against state-of-the-art methods. The results demonstrate that, under the initialization template setup, our algorithm excels in challenging scenarios and maintains robust performance as a tracking baseline.

\begin{table*}[t]
\centering
\caption{\textbf{Comparison with the state-of-the-art trackers on EgoTracks under VCT-VS and VCT-VC}. $\dagger$ indicates that this work was fine-tuned on EgoTracks.}
\label{tab:egotracks}
\resizebox{\textwidth}{!}{%
\begin{tabular}{@{}c|ccccccccc@{}}
\toprule
 &
  \multicolumn{5}{c|}{\textbf{VCT-VS}} &
  \multicolumn{4}{c}{\textbf{VCT-VC}} \\ \cmidrule(l){2-10} 
\multirow{-2}{*}{\textbf{Method}} &
  \textbf{AO$\uparrow$} &
  \textbf{F-score$\uparrow$} &
  \textbf{Precision$\uparrow$} &
  \textbf{Recall$\uparrow$} &
  \multicolumn{1}{c|}{\textbf{FPS$\uparrow$}} &
  \textbf{AO$\uparrow$} &
  \textbf{F-score$\uparrow$} &
  \textbf{Precision$\uparrow$} &
  \textbf{Recall$\uparrow$} \\ \midrule
GlobalTrack\cite{globaltrack} &
  23.63 &
  20.35 &
  31.28 &
  15.14 &
  \multicolumn{1}{c|}{6} &
  - &
  - &
  - &
  - \\
LTMU\cite{LTMU} &
  29.33 &
  27.46 &
  37.28 &
  21.74 &
  \multicolumn{1}{c|}{13} &
  - &
  - &
  - &
  - \\
ToMP\cite{tomp} &
  30.93 &
  20.95 &
  19.63 &
  22.46 &
  \multicolumn{1}{c|}{5} &
  - &
  - &
  - &
  - \\
ToMP ($\dagger$) &
  36.13 &
  28.11 &
  29.01 &
  27.26 &
  \multicolumn{1}{c|}{4} &
  37.32 &
  29.31 &
  30.22 &
  28.45 \\
MixFormer\cite{mixformer} &
  27.93 &
  25.54 &
  28.30 &
  23.27 &
  \multicolumn{1}{c|}{7} &
  28.63 &
  26.68 &
  29.55 &
  24.31 \\
SiamRCNN\cite{siamrcnn}($\dagger$) &
  45.37 &
  41.41 &
  56.11 &
  32.81 &
  \multicolumn{1}{c|}{4.5} &
  46.11 &
  43.20 &
  61.28 &
  33.36 \\
EgoSTARK\cite{egotracks}($\dagger$) &
  45.46 &
  43.33 &
  56.88 &
  34.99 &
  \multicolumn{1}{c|}{9} &
  47.33 &
  44.54 &
  59.86 &
  35.47 \\
\rowcolor[HTML]{C0C0C0} 
EgoHieraLoc(Ours) &
  45.99 &
  44.80 &
  58.11&
  36.45 &
  \multicolumn{1}{c|}{5} &
  49.91 &
  46.53 &
  62.57 &
  37.03 \\ \bottomrule
\end{tabular}%
}
\end{table*}
\section{Camera Pose Estimation and Alignment}
\label{app:camera}
Estimating camera poses from first-person perspective videos with significant viewpoint changes is a highly challenging task. EgoLoc\cite{egoloc} used a direct and efficient solution paradigm: COLMAP. As shown in Fig.\ref{fig:overall}(a), we also employ COLMAP \cite{colmap} for camera pose estimation and 3D reconstruction. By adjusting and optimizing the hyperparameter settings to better accommodate the characteristics of egocentric videos, we can obtain more valid camera poses in a shorter time compared to the COLMAP settings used in EgoLoc. 

We employ a keypoint matching strategy and Perspective-n-Point (PnP) solver for camera pose estimation. Overall, our approach comprises the following four steps. \textbf{First}, we estimate the camera intrinsics using Structure-from-Motion (SfM). \textbf{Second}, we extract and match keypoints from each frame in the video with keypoints extracted from Matterport3D scan panoramas. \textbf{Then}, leveraging the matched keypoints, we formulate and solve the PnP problem for each frame in the video to estimate the corresponding camera pose. \textbf{Finally}, we refine these poses using temporal constraints.

\noindent\textbf{Camera intrinsics prediction}
We use Laplacian transformation on non-blurry frames to construct a contiguous set of non-blurry images. To constrain the computational cost of SfM, we cap the number of selected frames at 25. The camera intrinsics are then estimated by running COLMAP automatic reconstruction model.

\noindent\textbf{Keypoint extraction and matching}
We employ SuperGlue\footnote{https://github.com/magicleap/SuperGluePretrainedNetwork} for keypoint extraction and matching. We begin by extracting keypoints from the scan panoramas $\{k_{\{p,n\}}, p\in P, n \in N_{kp}\}$, where $P$ denotes the number of panoramas and $N_{kp}$ represents the number of keypoints. Subsequently, we render RGB and depth images at each scan position, sweeping over pitch values $\in [-60,60]$ with a step size of 5 degrees and yaw values $\in [-180, 180]$ with a step size of 10 degrees. On average, we generate 9K images per scan. We extract keypoints from the video frames $\{k_{\{n,b\}}, n \in N_V, b \in B\}$, where $N_V$ signifies the number of images in the video and $B$ the number of keypoints. Once the keypoints are extracted, we iterate through each frame $n \in N_V$ in the video and match the extracted frame keypoints $\{k_{\{n,b\}},b \in B\}$, to all panorama keypoints $\{k_{\{w,r\}}, w \in W, r \in R\}$. We utilize pre-trained SuperPoint\footnote{https://github.com/rpautrat/SuperPoint} models for keypoint and descriptor extraction, and SuperGlue for the matching process.

\noindent\textbf{Initial pose estimation}
The positions of the 3D keypoints are computed using a pinhole camera model of the Matterport scan, in conjunction with the rendered panorama depth, camera intrinsics, and camera pose. The positions of the 2D keypoints are directly extracted from the video frame pixels. Subsequently, we employ the OpenCV library to solve the PnP problem, estimating the camera pose from the matched 3D and 2D point correspondences, utilizing the estimated camera intrinsics. Finally, we incorporate temporal constraints to increase the number of camera pose estimates.

\noindent\textbf{Temporal Refinement}. Following \cite{ego4d}, we optimize pose estimation by incorporating temporal constraints within the iterative process. Specifically, we begin by extracting 2D keypoints from the localization frame and matching them to non-localization frames within the video. We employ the same SuperGlue, along with utilizing the matched keypoints and the currently estimated pose. We triangulate novel 3D keypoints for the non-localization images. Subsequently, we solve a novel PnP configuration using these new keypoints. We iteratively apply this process until convergence.
\begin{lstlisting}[caption={Script code for model\_aligner}, label={lst:aligner}]
import subprocess
import shlex

def align_colmap_model(matterport_renderings_path):
    command = [
        'colmap',
        'model_aligner',
        '--ref_images_path', matterport_renderings_path,
        '--ref_is_gps', '0',
        '--robust_alignment', '1',
        '--alignment_type', 'custom',
        '--estimate_scale', '1',
        '--robust_alignment_max_error', '30'
    ]
    try:
        result = subprocess.run(
            command,
            check=True,
            capture_output=True,
            text=True
        )
        print("COLMAP output:\n", result.stdout)
    except subprocess.CalledProcessError as e:
        print(f"COLMAP failure: {e}")
\end{lstlisting}
\noindent\textbf{Alignment between Matterport Scan and Videos}. Since the poses provided by COLMAP are inconsistent with the Matterport world coordinate system, alignment is required. We employ a post-processing step, aligning the COLMAP reconstruction with the Matterport scan coordinate system via a Sim3 transformation. Specifically, we utilize COLMAP's \textit{model\_aligner} function and render at least five images with known camera poses from the Matterport scan. Subsequently, we estimate the Sim3 transformation between the COLMAP coordinate system and the Matterport scan coordinate system using the script in Listing.~\ref{lst:aligner}. This enables us to evaluate the results in the same coordinate system in which the annotators labeled the ground truth 3D bounding boxes.
\section{Additional Ablations}
\label{app:additionalablations}
\subsection{Impact of Training Data: VISOR Contribution Analysis}
\label{sec:data_ablation}
To isolate the architectural contribution from the effect of additional training data, 
we conduct an ablation study removing the VISOR dataset from the training pipeline. 
The model trained exclusively on Ego4D-VQ is compared against both our full model 
and the RELOCATE baseline (which uses only Ego4D-VQ training).
\begin{table}[t]
  \centering
  \caption{VISOR dataset ablation study on Ego4D-VQ2D and VQ3D validation sets.}
  \label{tab:visor_ablation_combined}
  
  \begin{minipage}[t]{0.49\textwidth}
    \centering
    \subcaption{Ego4D-VQ2D Validation Set}
    \label{tab:visor_ablation}
    \resizebox{\linewidth}{!}{%
      \begin{tabular}{lcccc}
        \toprule
        \textbf{Training Data} & \textbf{tAP25} & \textbf{stAP25} & \textbf{Rec.(\%)} & \textbf{Succ.(\%)} \\
        \midrule
        RELOCATE (Ego4D-VQ only) & 0.41 & 0.33 & 50.50 & 58.04 \\
        EgoHieraLoc w/o VISOR   & 0.41 & 0.33 & 50.88 & 60.45 \\
        ~~$\Delta$ vs. RELOCATE  & $+0.00$ & $+0.00$ & $+0.38\%$ & $\mathbf{+2.41\%}$ \\
        \midrule
        EgoHieraLoc w/ VISOR    & 0.43 & 0.35 & 51.52 & 62.30 \\
        ~~$\Delta$ vs. RELOCATE  & $+0.02$ & $+0.02$ & $+1.02\%$ & $\mathbf{+4.26\%}$ \\
        \bottomrule
      \end{tabular}%
    }
  \end{minipage}%
  \hfill 
  \begin{minipage}[t]{0.49\textwidth}
    \centering
    \subcaption{Ego4D-VQ3D Validation Set}
    \label{tab:visor_ablation_3d}
    \resizebox{\linewidth}{!}{%
      \begin{tabular}{lcccc}
        \toprule
        \textbf{Training Data} & \textbf{Succ.(\%)} & \textbf{Succ*(\%)} & \textbf{L2 ($\downarrow$)} & \textbf{QwP(\%)} \\
        \midrule
        EgoHieraLoc w/o VISOR & 80.12 & 97.89 & 1.38 & 84.48 \\
        EgoHieraLoc w/ VISOR  & 82.25 & 98.22 & 1.30 & 84.48 \\
        ~~Improvement         & $+2.13\%$ & $+0.33\%$ & $-0.08$ & $-$ \\
        \bottomrule
      \end{tabular}%
    }
  \end{minipage}
\end{table}

The following are the key findings: (i) Without VISOR, EgoHieraLoc achieves $60.45\%$ success rate vs. RELOCATE's $58.04\%$, demonstrating a $\mathbf{+2.41\%}$ improvement purely from our three-module design (DPM, QAM, RAM) and GSJC weighting. (ii)The additional $1.85\%$ gain ($62.30\% - 60.45\%$) comes from leveraging VISOR's dense pixel-level supervision during $30\%$ of training iterations, providing supplementary mask guidance that refines the segmentation branch. (iii) This decomposition clarifies that the core architectural innovations contribute more to performance improvement than the additional training data, validating the design choices of DPM, QAM, and RAM over simpler feature-matching baselines. We further evaluate this variant on Ego4D-VQ3D (Table.\ref{tab:visor_ablation_3d}), which demonstrates that the consistent improvement across both 2D and 3D tasks confirms that VISOR provides complementary mask supervision that enhances the robustness of our segmentation pipeline.

\subsection{Impact of components of QAM on EgoTracks}
\label{app:qamegotracks}
\begin{figure*}[t]
\centering
\includegraphics[width=\textwidth]{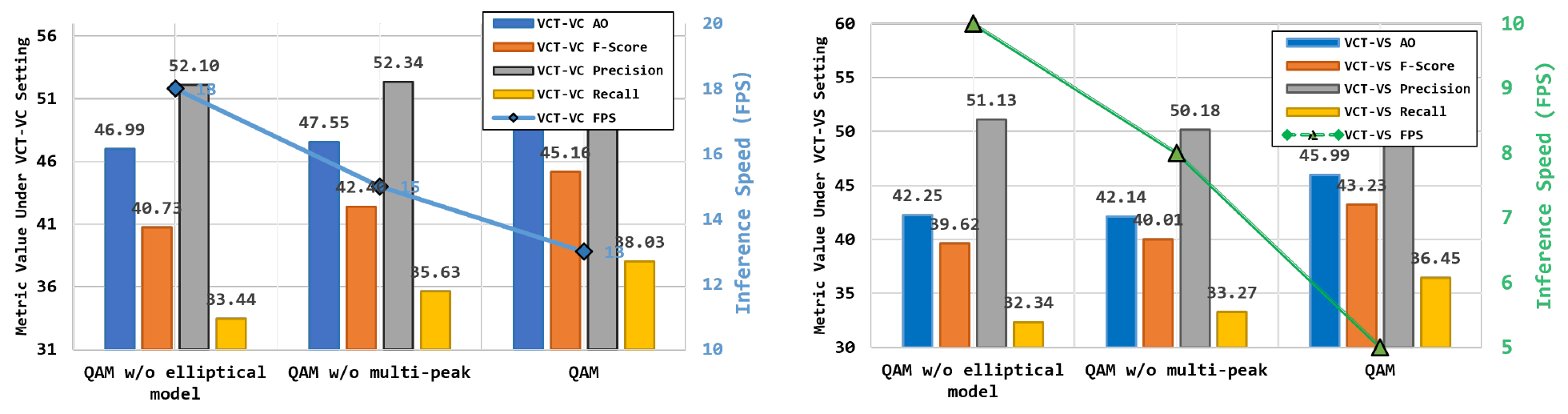}
\caption{\textbf{Ablations for QAM modules on EgoTracks.}}
\label{app:qamegotracksablations}
\end{figure*}

In addition, we analyze the critical components of QAM on EgoTracks. Two variants of QAM are adopted, consistent with the experiments on Ego4D-VQ: (i) QAM without elliptical model constraints and (ii) QAM without multi-peak detection. As illustrated in Fig.\ref{app:qamegotracksablations}, the absence of the elliptical model in the VCT-VC configuration results in a decrease in F-score from 45.16 to 40.73 and recall from 38.03 to 33.44. Removing the multi-peak component reduces the F-score from 45.16 to 42.40, recall from 38.03 to 35.63, and precision from 55.57 to 52.34. Under the VCT-VS setting, the lack of elliptical constraints causes recall to drop from 36.45 to 32.34 and F-score from 43.23 to 39.62. Similarly, removing multi-peak detection leads to a reduction in F-score from 43.23 to 40.01, recall from 36.45 to 33.27, and precision from 53.11 to 50.18. The experiments on elliptical model constraints indicate an asymmetric impact on recall, demonstrating that recall is more sensitive to missed detections than precision. When the target undergoes deformation, the DCF loses confidence in the spatial extent of the target, causing tracking failure rates to increase significantly. Elliptical constraints effectively encode continuous geometric priors into the target region, providing a structurally grounded reference for the tracker. In the experiments regarding multi-peak detection, it is observed that unresolved peak ambiguities cause the tracker to lock onto distractor responses, creating false associations that directly degrade detection-level precision. Significant differences in runtime exist between the two configurations. In the VCT-VS setting, the tracker processes every frame in causal order starting from the first frame, representing a single unidirectional scan of the entire temporal sequence. In the VCT-VC setting, the tracker executes twice for the same video by running forward and backward from the visual crop frame, which doubles the total number of frames processed per video segment. Although the forward and backward runs in VCT-VC only cover portions of the video timeline anchored by the visual crop, VCT-VS must handle the full duration of the video in a single uninterrupted inference process. This accumulates longer total inference sequences and subsequently lowers the effective frame rate. Furthermore, VCT-VS lacks the visual proximity advantage inherent in VCT-VC, where high-quality visual crop templates serve as initialization references. In VCT-VC, the initial frames exhibit high visual similarity to the template, a structural proximity that reduces costly re-detection and template update operations triggered by appearance drift and indirectly increases throughput. In contrast, VCT-VS must cope with the full range of appearance changes accumulated throughout the entire video from the very first frame. Ablation results demonstrate that the complete QAM module maximizes tracking robustness by integrating geometric and distribution priors, yielding the highest F-score and average overlap (AO) metrics. Specifically, elliptical modeling serves as a vital supplement to the underlying filter by providing critical spatial morphology priors that adapt to non-rigid deformations. Meanwhile, multi-peak detection explicitly addresses localization ambiguities that arise when the target response map exhibits multiple discontinuous peaks. The removal of either component deprives the tracker of essential shape awareness and peak discrimination capabilities, leading to tracking failures during severe deformations. However, the computation of these advanced spatial-visual constraints introduces unavoidable processing overhead. Consequently, integrating the full QAM results in a decrease in inference speed from 18 FPS to 13 FPS in the VCT-VC setting and from 10 FPS to 5 FPS in the VCT-VS setting.
\subsection{Complete Analysis of GSJC}
\label{app:gsjc}
As detailed in Table~.\ref{tab:aggregation}, the experimental results highlight the critical role of semantic weighting factors in improving localization accuracy within the VQ3D task. Among the individual components of $\vartheta^{sem}$, $\mathcal{P}_{max}$ yields the most significant single-factor contribution, raising the success rate from 78.13\% to 79.12\% and reducing the Angle error by 0.12. The integration of all three semantic metrics ($\mathcal{P}_{ave}$, $\mathcal{P}_{thr}$, and $\mathcal{P}_{max}$) further boosts the success rate to 81.42\% and decreases the L2 error from 1.66 to 1.37. This trend suggests that combining diverse semantic evidence effectively filters out low-quality frames, providing a more robust foundation for spatial reasoning than relying on any single metric. The introduction of geometric reliability factors as additional weights provides incremental performance gains over the semantic-only baseline. Among the geometric components, $\varpi^{reproj}$ proves to be the most influential factor, achieving a Succ score of 81.82\% and a Succ* of 98.11\% when added to the full semantic set. In comparison, $\varpi^{depth}$ and $\varpi^{tri}$ yield slightly lower but consistent improvements. These metrics physically constrain the 3D predictions by assessing reprojection consistency and triangulation geometry, ensuring that the aggregation process prioritizes viewpoints with higher geometric fidelity. The complete integration of both semantic and geometric components in the GSJC scheme yields the optimal performance across all metrics, achieving a Succ of 82.25\% and an L2 error of 1.30. This represents a cumulative improvement of 4.12\% in Succ and a 21.7\% reduction in L2 error relative to the baseline. Notably, while accuracy and angular metrics show consistent optimization as more components are added, the QwP remains constant at 84.48\%. Following the metric definition in \cite{ego4d}, this is by design, not coincidence: QwP measures the fraction of queries with available camera poses for both response-track frames and the query frame. This condition depends solely on the success of offline COLMAP camera-pose estimation on raw video sequences. Since this pose estimation is a fixed pre-processing step 
performed prior to and independently of any network component, QwP depends only on 
the intrinsic geometric and photometric properties of the videos rather than on the model. Consequently, architectural or 
loss-function ablations—which alter how accurately an object is localized 
but not whether a camera pose can be recovered—leave both the numerator and 
denominator of QwP unchanged. In contrast, the Success and Success$^{*}$ rates, 
which depend on the predicted 3D location's accuracy, vary meaningfully 
across variants and thus serve as the primary indicators of each component's 
contribution.

In the original formulation, GSJC weights are multiplied and normalized as a weighted average.
To ensure numerical stability when all weights are extremely small, we add a small epsilon and a fallback strategy.
The total weight becomes
\[
\tilde{\vartheta}_i=\vartheta^{\text{sem}}_i\cdot \varpi^{\text{depth}}_i\cdot \varpi^{\text{reproj}}_i\cdot \varpi^{\text{tri}}_i,
\quad \tilde{\vartheta}_i \leftarrow \max(\tilde{\vartheta}_i,\epsilon).
\]
Then the multi-view aggregation is
\[
A\left(\{(x_j,y_j,z_j,\tilde{\vartheta}_j)\}_{j=1}^{N_C}\right)=
\frac{\sum_{j=1}^{N_C} \tilde{\vartheta}_j \,[x_j,y_j,z_j]}
{\sum_{j=1}^{N_C} \tilde{\vartheta}_j + \epsilon}.
\]
If $\sum_{j=1}^{N_C} \tilde{\vartheta}_j < \epsilon$, we fall back to uniform averaging.
\begin{figure}[]
\centering
\includegraphics[width=\textwidth]{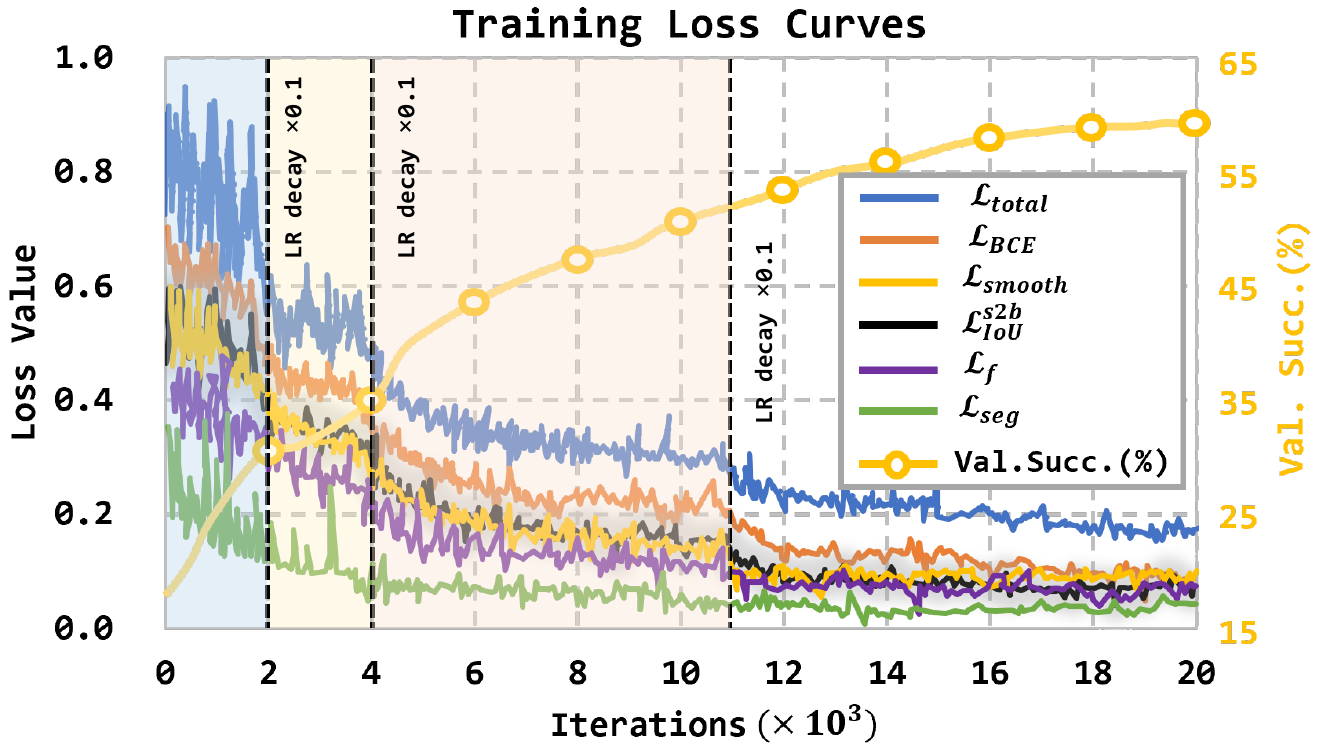}
\caption{\textbf{Training losses of EgoHieraLoc.} We plot the raw mini-batch loss curves to illustrate convergence behavior under a multi-step learning rate scheduler. Vertical dashed lines denote 0.1 learning rate decay. Notably, the success rate (yellow curve, right y-axis) on validation set exhibits a pronounced acceleration following each annealing stage.}
\label{fig:trainingloss}
\end{figure}
\subsection{Training Dynamics and Loss Component Analysis}
\label{app:trainingloss}
Fig.~\ref{fig:trainingloss} presents the complete training dynamics of EgoHieraLoc over 20,000 iterations. All loss functions decrease in expectation, and the success rate (Val. Succ., right axis) rises steadily from initialization to a plateau of 62.30\%, confirming stable convergence without overfitting. The composite loss $\mathcal{L}_{total}$ descends from the initial stage to approximately 0.21 at convergence, representing a reduction of 77.2\%. Specifically, the learning rate schedule partitions training into three distinct optimization phases, each exhibiting characteristic loss behaviors. During the rapid feature acquisition phase (0 to 2,000 iterations), a linear warm-up scales the learning rate from 0 to a peak of $1.5\times10^{-3}$ within the first 250 iterations, followed by full-rate optimization until the first step decay at 2,000. In this phase, $\mathcal{L}_{\text{BCE}}$ and $\mathcal{L}_{\text{smooth}}$ experience the most dramatic absolute declines, as the model rapidly acquires basic frame-query discrimination and coarse spatial regression capabilities. High-frequency oscillations in this period stem from (i) stochastic mini-batch sampling across heterogeneous data sources including Ego4D-VQ and VISOR, (ii) the inherent non-stationarity of the AdamW optimizer at high learning rates, and (iii) competing gradient signals from five simultaneous loss terms shaping the shared feature representation. The DCF loss $\mathcal{L}_f$ exhibits the highest initial amplitude and the fastest early decline. We emphasize that this decline does \emph{not} arise from iteratively optimizing the correlation filter itself: for each query, the filter $f^{*}$ is obtained analytically through the closed-form frequency-domain solution. Instead, $\mathcal{L}_f$ measures the residual of this closed-form filter evaluated on the end-to-end fine-tuned backbone features $\mathcal{F}(\cdot)$, so its steep early drop indicates that the shared backbone rapidly learns representations under which the analytically computed filter already yields a sharp Gaussian-like response. The segmentation-geometry co-refinement phase (2,000 to 11,000 iterations) begins with the first decay ($\times0.1$ at 2,000), which lowers the learning rate to $1.5\times10^{-4}$ and immediately dampens oscillation across all curves. The most significant relative declines occur in $\mathcal{L}_{\text{IoU}}^{s2b}$, which drops 44\% relative to its value at 2,000, and $\mathcal{L}_{\text{seg}}$, which falls 51\%, indicating that mask-to-box alignment and mixed segmentation supervision benefit from the stable feature representations established in the first stage. Notably, $\mathcal{L}_{\text{seg}}$ exhibits a distinctive two-stage descent pattern within this phase: a steep drop from 2,000 to 5,000 iterations followed by a shallower but sustained decline from 5,000 to 11,000. This trend reflects the mixed supervision strategy where the initial steep descent corresponds to the model fitting densely supervised VISOR samples that account for 30\% of iterations and provide full pixel-level guidance. The subsequent gradual descent indicates progressive adaptation to weakly supervised Ego4D-VQ pseudo-masks which comprise 70\% of iterations with a coefficient of $\lambda_{\text{pdo}} = 0.3\eta_4$, where noisy labels from SAM constrain the rate of improvement. This observation validates the necessity of asymmetric loss weighting, as the noisy Ego4D labels would otherwise dominate $\mathcal{L}_{\text{seg}}$ and impede convergence. A second step decay ($\times0.1$ at 4,000) further reduces gradient noise, after which the validation success rate experiences its most rapid rise, increasing by approximately 11.3 percentage points between 4,000 and 11,000 iterations. This suggests that the mid-training stage is critical for the model to internalize discriminative query-video correspondences. The final step decay ($\times0.1$ at 11,000) initiates a precision refinement phase with low variance. All five losses stabilize within tight plateaus, where Smooth-L1 regression retains residual error on geometrically ambiguous samples while $\mathcal{L}_f$ settles to a low, stable residual, indicating that the closed-form filter consistently produces sharp responses on the now-converged backbone features. The success rate reaches 62.30\%, confirming that the model effectively extracts learnable signals from the training data without overfitting. We identify three interaction patterns. First, the synchronous decline of $\mathcal{L}_{\text{IoU}}^{s2b}$ and $\mathcal{L}_{\text{seg}}$ indicates that mask quality and box tightness are learned jointly; improvements in the segmentation mask directly reduce background pixels within the fitted box, represented by lower $\mathcal{N}_{IS}^-$, thereby decreasing $\mathcal{L}_{\text{IoU}}^{s2b}$ and confirming that both losses provide complementary signals. Second, $\mathcal{L}_{\text{smooth}}$ maintains the highest absolute value throughout training, reflecting its $\eta_2 = 0.4$ weight and the intrinsic difficulty of bounding box coordinate regression under highly variable egocentric viewpoints. The persistently higher residual of $\mathcal{L}_{\text{smooth}}$ relative to $\mathcal{L}_{\text{IoU}}^{s2b}$ suggests that precise coordinate localization is a more challenging sub-task than holistic mask-box alignment, matching the observation that tAP25 lags behind success rate in the main results. Third, $\mathcal{L}_f$ reaches its stable residual, around 3,000 iterations. Since the filter is solved in closed form rather than optimized, this early stabilization reflects that the backbone features feeding the filter become discriminative faster than the segmentation and box-regression branches mature. This rapid convergence provides a stable spatial prior to the QAM module, allowing it to guide RAM refinement reliably before the segmentation branch matures.
\subsection{Hyperparameter Sensitivity Analysis for QAM}
\label{sec:hyperparams_sensitivity}
We evaluate the robustness of our method against variations in five key hyperparameters 
across $\pm 30\%$ ranges around the selected optimal values. This analysis assesses 
whether our design is overly tuned to the validation set. Across all five hyperparameters, performance fluctuations within $\pm 30\%$ perturbations range from $\pm 0.5\%$ to $\pm 1.1\%$, demonstrating \textit{moderate robustness} to hyperparameter choices. The selected hyperparameters ($\alpha_{QAM}=1.0, \gamma_{QAM}=0.4, \eta_{QAM}=0.7, \lambda_e=0.5$) were optimized on the Ego4D-VQ validation set through grid search. For deployment on new datasets (e.g., EgoTracks, TREK-150), we recommend re-tuning these five parameters using the same grid-search protocol to maintain performance. The minimal degradation between validation and test performance (e.g., tAP25: $0.43 \to 0.44$) suggests that hyperparameter overfitting is not a major concern.
\begin{table}[htbp]
\centering
\caption{\textbf{Hyperparameter sensitivity analysis for QAM on Ego4D-VQ2D validation set.} Optimal values are marked with \ding{51}. Results show mean $\pm$ std over 3 runs.}
\label{tab:hyperparam_sensitivity_grid}
\resizebox{\columnwidth}{!}{%
\begin{tabular}{lcccc | lcccc}
\toprule
\textbf{Value} & \textbf{tAP25} & \textbf{stAP25} & \textbf{Rec.} & \textbf{Succ.} & \textbf{Value} & \textbf{tAP25} & \textbf{stAP25} & \textbf{Rec.} & \textbf{Succ.} \\
\midrule
\multicolumn{5}{c|}{\textit{(a) Threshold Scale ($\alpha_{\text{QAM}}$)}} & \multicolumn{5}{c}{\textit{(b) Deformation Decay ($\gamma_{\text{QAM}}$)}} \\
0.5 ($-30\%$) & 0.41 & 0.33 & 50.2 & 61.3 & 0.2 ($-50\%$) & 0.42 & 0.34 & 50.6 & 61.7 \\
\textbf{1.0} \ding{51} (Default) & \textbf{0.43} & \textbf{0.35} & \textbf{51.5} & \textbf{62.3} & \textbf{0.4} \ding{51} (Default) & \textbf{0.43} & \textbf{0.35} & \textbf{51.5} & \textbf{62.3} \\
1.5 ($+30\%$) & 0.42 & 0.34 & 50.8 & 61.8 & 0.6 ($+50\%$) & 0.41 & 0.32 & 50.2 & 61.4 \\
\midrule
\multicolumn{5}{c|}{\textit{(c) Ellipse Smoothing ($\eta_{\text{QAM}}$)}} & \multicolumn{5}{c}{\textit{(d) Ellipse Weight ($\lambda_e$)}} \\
0.5 ($-30\%$) & 0.42 & 0.34 & 51.0 & 62.0 & 0.3 ($-40\%$) & 0.41 & 0.32 & 50.1 & 61.5 \\
\textbf{0.7} \ding{51} (Default) & \textbf{0.43} & \textbf{0.35} & \textbf{51.5} & \textbf{62.3} & \textbf{0.5} \ding{51} (Default) & \textbf{0.43} & \textbf{0.35} & \textbf{51.5} & \textbf{62.3} \\
0.9 ($+30\%$) & 0.41 & 0.33 & 50.4 & 61.9 & 0.7 ($+40\%$) & 0.42 & 0.34 & 50.9 & 62.1 \\
\bottomrule
\end{tabular}%
}
\end{table}
\begin{figure}[t!]
\centering
\includegraphics[width=\textwidth]{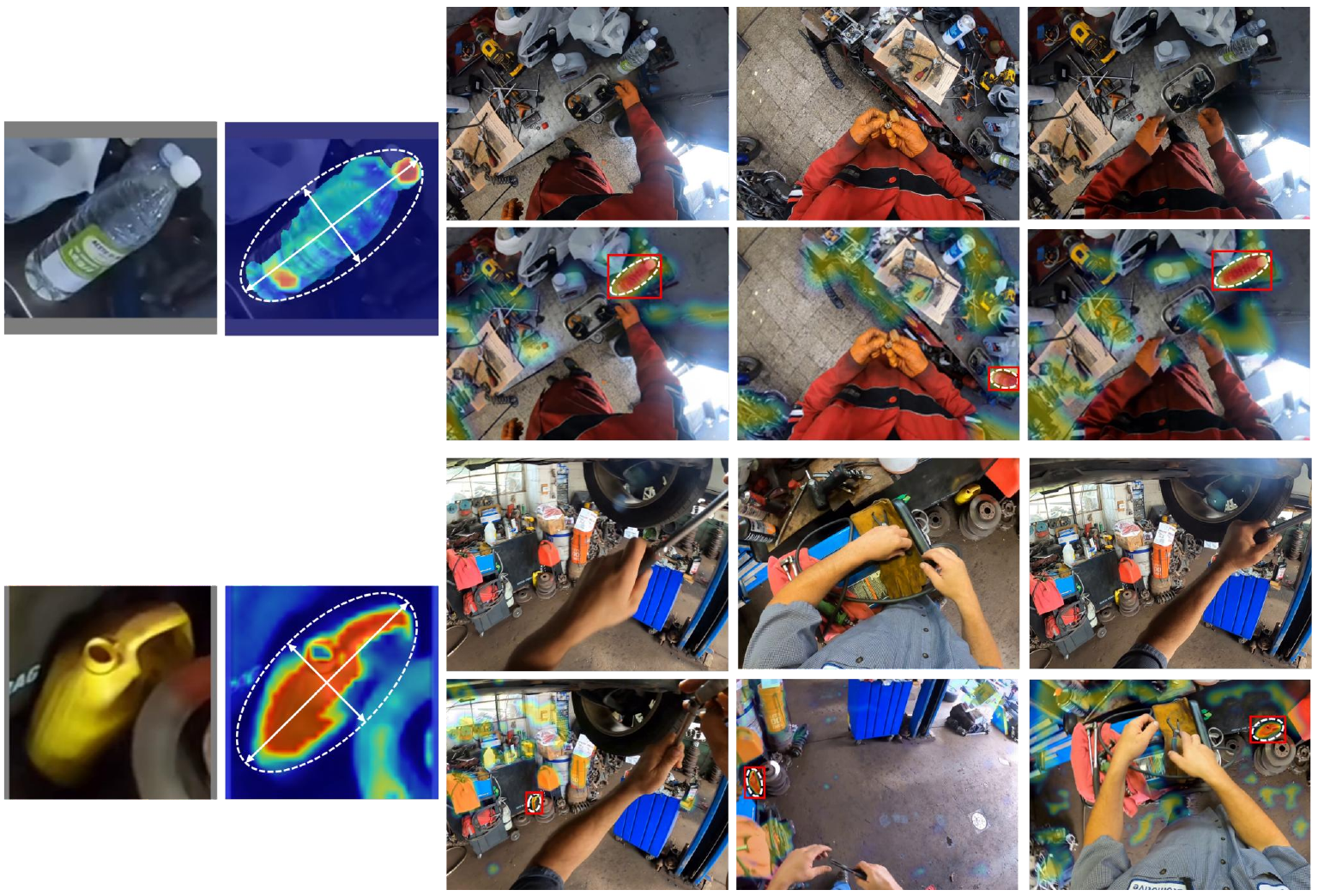}
\caption{
\textbf{Visualization of the dynamic elliptical constraint in QAM.} 
\textbf{Left two columns:} the query image (left) and its corresponding 
adaptive elliptical fitting overlaid on the response map (right), where the 
white dashed ellipse and its semi-axes ($a$, $b$) are automatically derived 
from the query's bounding-box dimensions (Eq. (\ref{eq:ellipse_update})). 
For the elongated bottle (top), the constraint yields a high-aspect-ratio 
ellipse ($a \gg b$), whereas for the compact power tool (bottom), it produces 
a near-circular ellipse ($a \approx b$). 
\textbf{Right ($2\times3$) grid:} for each example, the \textit{top row} shows 
three representative frames from the egocentric video sequence, and the 
\textit{bottom row} shows the corresponding localization response maps produced 
by EgoHieraLoc. The red dashed ellipse indicates the anisotropic spatial prior applied 
during peak refinement (Eq. (\ref{eq:distance_metric})). The elliptical 
constraint suppresses spurious off-axis peaks (e.g., background clutter, 
similar distractors) and concentrates the response energy along the object's 
principal orientation, yielding accurate and stable localization even under 
severe viewpoint changes and partial occlusion.}
\label{fig:ellipse_vis}
\end{figure}
\subsection{Visual Analysis of the Elliptical Constraint in QAM}
\label{app:ellipse_visualization}
To provide an intuitive understanding of how the dynamic elliptical constraint contributes to robust localization, we visualize the fitted ellipses and their 
effect on target retrieval across diverse egocentric scenes in 
Fig.\ref{fig:ellipse_vis}. As shown in the left columns of Fig.\ref{fig:ellipse_vis}, the elliptical 
constraint automatically adapts its geometry to the intrinsic shape of the 
query. The elongated water bottle induces a slender ellipse with a large 
semi-major-to-semi-minor axis ratio ($a/b \approx 3.5$), reflecting its 
vertical elongation. In contrast, the compact power tool produces a 
near-isotropic ellipse ($a/b \approx 1.3$). This shape-awareness is crucial: 
a fixed isotropic (circular) prior would either over-penalize the elongated 
object's valid extent along its major axis or under-constrain the compact 
object, both leading to localization drift. The bottom rows of each example demonstrate how the elliptical constraint 
reshapes the raw DCF response. Without the constraint, the response maps 
exhibit multiple competing peaks—particularly in cluttered workshop scenes 
where reflective surfaces, tools, and hands generate strong distractor 
responses. The anisotropic distance metric in Eq.(\ref{eq:distance_metric}) 
penalizes candidate peaks that deviate from the object's principal axes, 
effectively down-weighting off-axis distractors. As a result, the refined 
response (bottom rows) concentrates sharply around the true target, with 
the red dashed ellipse tightly enclosing the correct object. Across the three frames of each sequence, the query undergoes substantial 
scale and perspective changes induced by head-mounted camera motion. The 
recursive smoothing of the ellipse semi-axes (Eq.(\ref{eq:ellipse_update}), 
$\eta_{\text{QAM}} = 0.7$) allows the constraint to gradually adapt to these 
variations while maintaining temporal stability. Notably, even when the 
target is partially occluded (e.g., the bottle held against a dark background, 
top-right frame), the elliptical prior—combined with the cumulative 
deformation field $\Omega^{(v_i)}$—preserves accurate localization by 
leveraging the object's expected spatial extent and motion trajectory. These qualitative observations align with the ablation results in Fig. 7 of 
the main paper, where removing the elliptical constraint degrades the VQ2D 
success rate from $62.30\%$ to $60.39\%$ ($-1.91\%$). The visualization here 
reveals the underlying mechanism: the constraint's primary benefit lies in 
distractor suppression and shape-consistent peak selection, which are 
especially valuable in the visually complex, distractor-rich environments 
characteristic of egocentric footage.
\section{Adaptive Fallback for SAM Failures}
\label{app:qamfailure}
\begin{figure}[t]
  \centering
  \begin{minipage}[t]{0.47\textwidth}
    \centering
    \includegraphics[width=\textwidth]{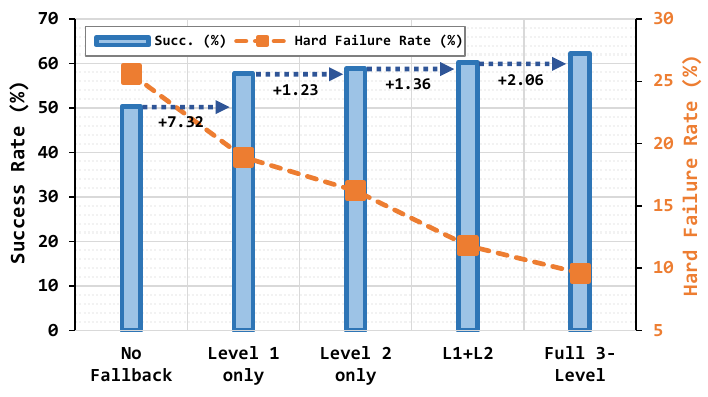}
    \centerline{(a)}
  \end{minipage}
  \hfill
  \begin{minipage}[t]{0.47\textwidth}
    \centering
    \includegraphics[width=\textwidth]{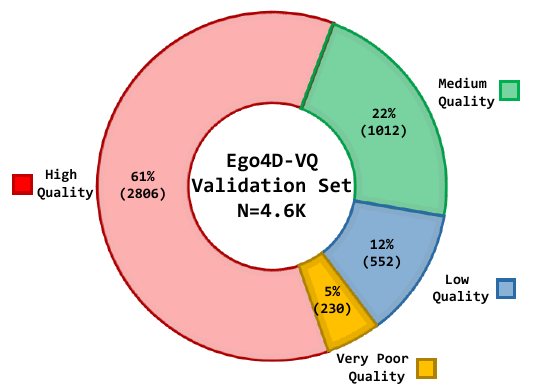}
    \centerline{(b)}
  \end{minipage}
  \caption{\textbf{Analysis of the adaptive fallback mechanism.} (a) Quantitative ablation of different fallback configurations. The proposed ``Full 3-Level'' strategy effectively mitigates catastrophic collapses, maximizing the success rate while significantly reducing the hard failure rate compared to the ``No Fallback'' baseline. (b) The distribution of SAM confidence scores ($s_{\text{SAM}}$) on the validation set, which validates the necessity of the fallback mechanism and guides the empirical thresholding.}
  \label{fig:combined_degradation}
\end{figure}

Visual queries cropped from video streams often lack sharp contours or sufficient discriminability. While the initial stage relies on the strong segmentation priors of the Segment Anything Model (SAM), erroneous masks can introduce severe interference during the subsequent retrieval phase. Because the DPM depends heavily on precise foreground and background feature sets, $M_O$ and $M_B$, inaccurate masks cause the DPM to extract inverted discriminative features. This inversion triggers a catastrophic collapse of the localization system. To detect these failures without requiring additional annotations, the predicted IoU score $s_{\text{SAM}} \in [0,1]$ from SAM is utilized as an unsupervised quality gating signal. Following a success-recall calibration on the validation set, an empirical threshold of $\tau_{\text{SAM}} = 0.6$ is established to trigger an adaptive fallback mechanism. Specifically, rather than employing a binary rejection strategy, a three-level degradation mechanism is designed to handle varying degrees of mask failure. In \textbf{Level 0}, which represents the normal operation with $s_{\text{SAM}} \ge 0.6$, the system uses the original SAM masks to construct $M_O$ and $M_B$, allowing the DPM to exert full background suppression. In \textbf{Level 1}, where the mask is only partially reliable ($0.4 \le s_{\text{SAM}} < 0.6$), a confidence-weighted soft mask is introduced, defined as $\tilde{S}(Q) = s_{\text{SAM}} \cdot S(Q) + (1 - s_{\text{SAM}}) \cdot \mathbf{1}$. This strategy smoothly expands the foreground sampling range to prevent the loss of target features caused by excessive mask contraction. In \textbf{Level 2}, where the mask is deemed entirely untrustworthy ($s_{\text{SAM}} < 0.4$), $M_O$ is set to include the global crop features, and $M_B$ is defined as $\emptyset$. Crucially, under this extreme fallback mode, a weight gain of $\beta_{\text{QAM}} = 1.5$ is applied to the Query-Aware Module (QAM). This leverages the frequency-domain robustness of the QAM, which remains independent of spatial masks, to guide the spatial localization. As shown in Fig.~\ref{fig:combined_degradation}, we validate these configurations. The "No Fallback" baseline, which rigidly applies Level 0 logic to all queries regardless of mask quality, suffers from a high hard failure rate. Equipping the system with individual fallback strategies ("Level 1 only" or "Level 2 only") yields partial improvements. Ultimately, the "Full 3-Level" strategy, which dynamically routes queries across Level 0, 1, and 2 based on $s_{\text{SAM}}$, achieves the optimal balance, maximizing the success rate while significantly reducing hard failures.
\section{Failure Case Analysis}
\label{app:failure}
\begin{figure*}[t]
\centering
\includegraphics[width=\textwidth]{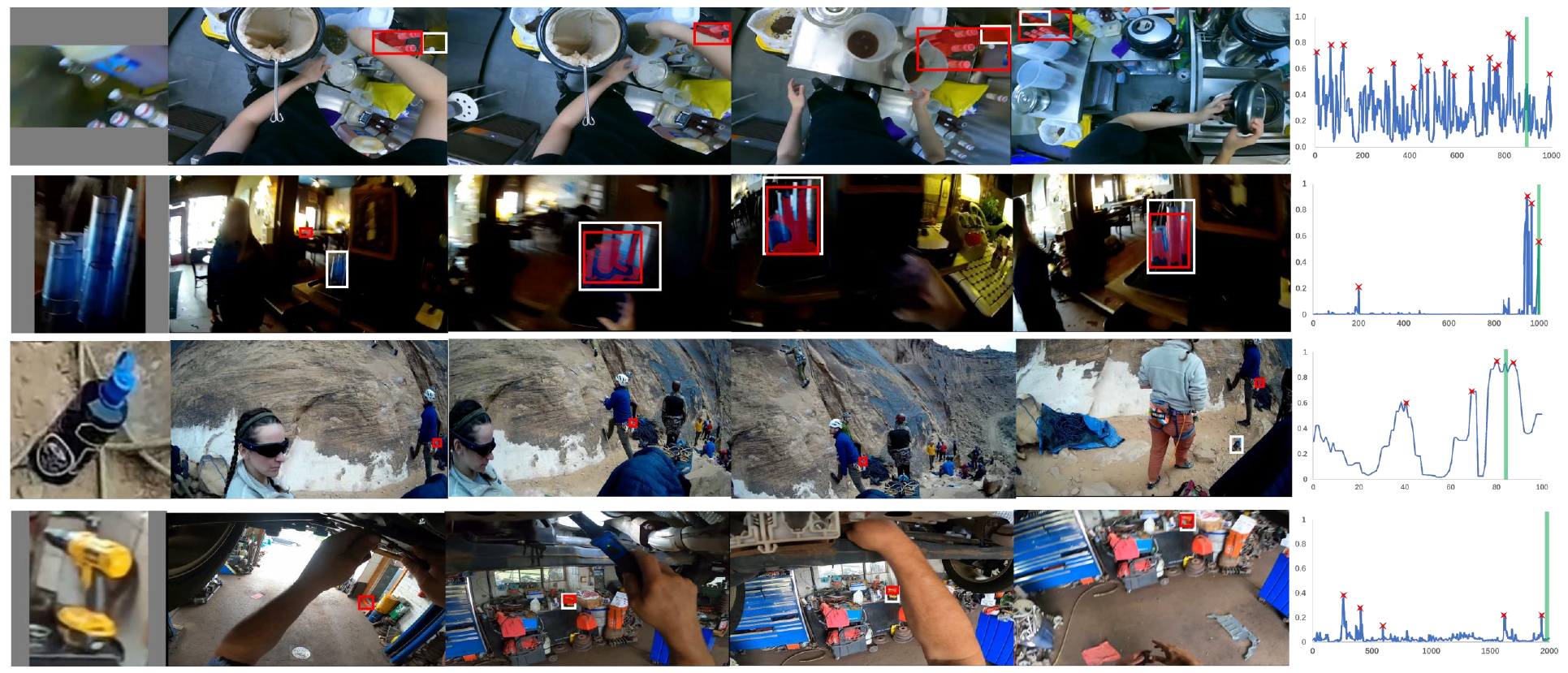}
\caption{\textbf{Failure analysis in Ego4D-VQ2D}. The left column displays the visual queries, and each corresponding row on the right presents the video response tracks. Within these trajectories, the white boxes denote the ground truth, while the red masks and their associated red bounding boxes represent the predictions generated by EgoHieraLoc.}
\label{fig:failureanalysis}
\end{figure*}

As illustrated in Fig.~\ref{fig:failureanalysis}, while the proposed quality-aware fallback mechanism successfully mitigates a large portion of segmentation errors, certain extreme scenarios in the Ego4D-VQ2D dataset remain unresolved. These residual failure cases reveal the fundamental limitations of appearance-based matching under severe egocentric visual degradation. The 1st two rows present instances of extreme feature-level ambiguity coupled with dense identical distractors. In the first row, severe motion blur physically destroys the high-frequency texture of the query, while the environment is cluttered with identical objects (bottles and caps). In the 2nd row, the target (blue cups) exhibits transparency and drastic viewpoint deformations. In such scenarios, even when the system gracefully degrades to the frequency-domain Query-Aware Module (QAM), the fundamental lack of discriminative texture renders the fallback mechanism ineffective. Consequently, the temporal response curves (right column) become highly noisy, and the model erroneously locks onto visually similar distractors. The 3rd row highlights the challenges associated with extreme scale variations and background camouflage. The query object is exceptionally small and shares an almost identical color distribution with the rugged terrain. The scarcity of high-resolution feature representations causes the model to fail in differentiating the true target from the background noise, demonstrating that current multi-scale feature pyramids still struggle to isolate sub-pixel level targets against highly textured environments. The 4th row demonstrates failures caused by dynamic occlusion and extreme egocentric viewpoint drifts. As the camera wearer interacts with the environment (e.g., repairing machinery), the target (a power drill) undergoes drastic geometric deformation and is heavily occluded by the user's hands. The algorithm struggles to maintain temporal coherence when the target's appearance is severely fragmented. These failure modes suggest that relying solely on 2D appearance modeling and spatial correlation has its upper bounds in unconstrained egocentric vision. Future research must extend beyond appearance features by integrating explicit 3D geometric priors, temporal motion kinematics, and hand-object interaction contexts to ensure robustness against absolute visual degradation.
\begin{figure*}[!h]
\centering
\includegraphics[width=0.95\textwidth]{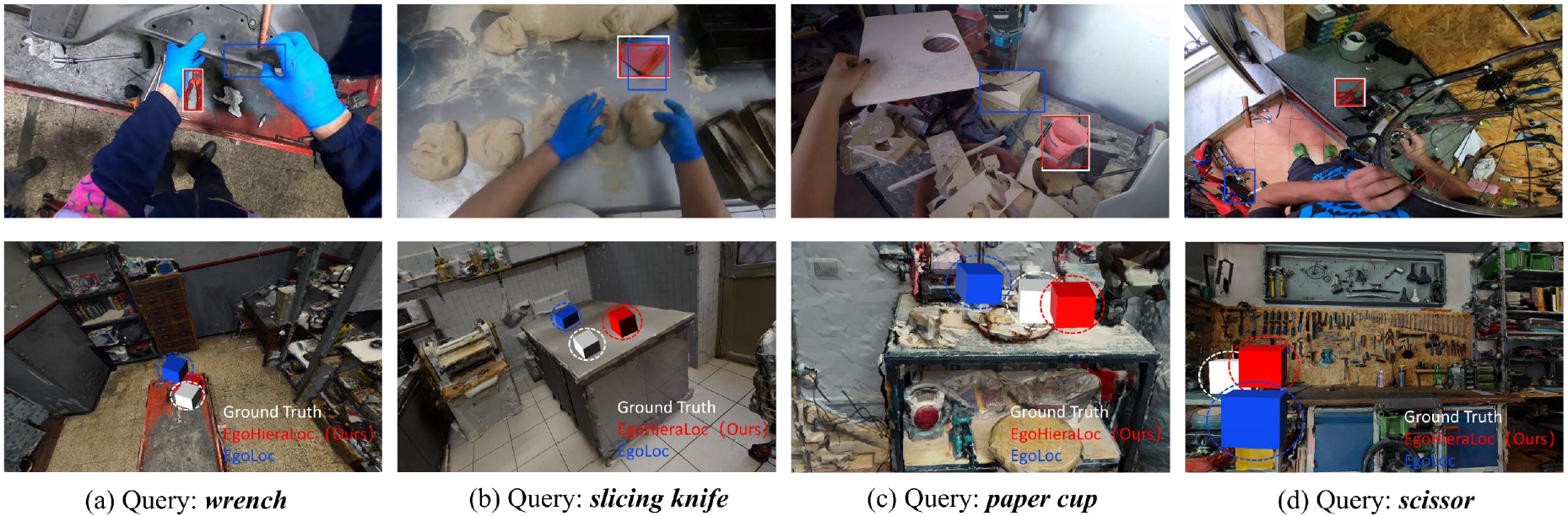}
\caption{\textbf{Visualization of back-projected 2D–to–3D.} We compare back-projected 3D locations of EgoHieraLoc against EgoLoc and Ground-Truth (GT). While both methods succeed in simple cases (a-b), EgoLoc suffers large displacement errors in challenging scenarios (c-d) due to inaccurate 2D responses. In contrast, EgoHieraLoc maintains robustness and closely matches the GT. Note: Visualizations use predicted 3D centers with GT box dimensions/orientation, as these are unknown during inference.}
\label{fig:vis3d}
\end{figure*}
\section{2D-to-3D Visualization}
\label{app:vis}
Fig.~\ref{fig:vis3d} presents a qualitative comparison of back-projected 3D object positions estimated by EgoHieraLoc (red) and EgoLoc (blue) against the ground truth (white) across four query cases in Matterport scanns. For each case, the top row shows the 2D response in the retrieved video frame, where the image-space centroid determined by the predicted bounding box is subsequently lifted to 3D space via depth back-projection. The bottom row renders the final 3D centroid predictions and the ground truth bounding box in the scene coordinate system. The scale and orientation of the ground truth box are fixed to their actual values across all methods since these quantities are inaccessible during inference, thus serving as a shared controlled variable. For query (a), the 2D bounding boxes retrieved by the two methods show noticeable differences. Conversely, for query (b), both methods achieve accurate 2D localization with their back-projected 3D centroids aligning closely with the ground truth. This suggests that when the object appearance is sufficiently discriminative and the scene context lacks ambiguity, both approaches can provide reliable 3D placement estimation. However, any deviation in 2D localization is significantly amplified when back-projected into 3D space. In the scene corresponding to query (c), the target is embedded in a visually cluttered desktop environment containing multiple objects with similar color and texture statistics. EgoLoc produces an inaccurate 2D response biased toward distractor regions. This bias introduces a large lateral displacement error in the back-projected 3D estimation, which is evident from the substantial spatial offset of the blue box relative to the ground truth. In contrast, EgoHieraLoc utilizes hierarchical discriminative parsing to suppress background interference, generating a target-centric 2D prediction and a 3D estimation that closely matches the actual position. For query (d), the object is small and observed from a top-down egocentric viewpoint, severely compressing its distinct geometric features. EgoLoc again fails to produce reliable 2D localization, resulting in a substantial distance error in the 3D projection. EgoHieraLoc mitigates this failure through frequency-domain correlation operations within the query-aware module. This mechanism maintains the target identity under severe viewpoint changes and yields a 3D centroid prediction aligned with the ground truth within the acceptable localization tolerance. These qualitative visualizations demonstrate that inaccurate 2D responses are a primary source of 3D localization errors. This finding validates the core design priority of EgoHieraLoc, which explicitly dedicates representational capacity to robust 2D spatial localization as a fundamental prerequisite for accurate 3D object placement in egocentric video understanding.
\begin{figure}[htbp]
  \centering
  \begin{minipage}[t]{0.48\textwidth}
    \centering
    \includegraphics[width=\linewidth]{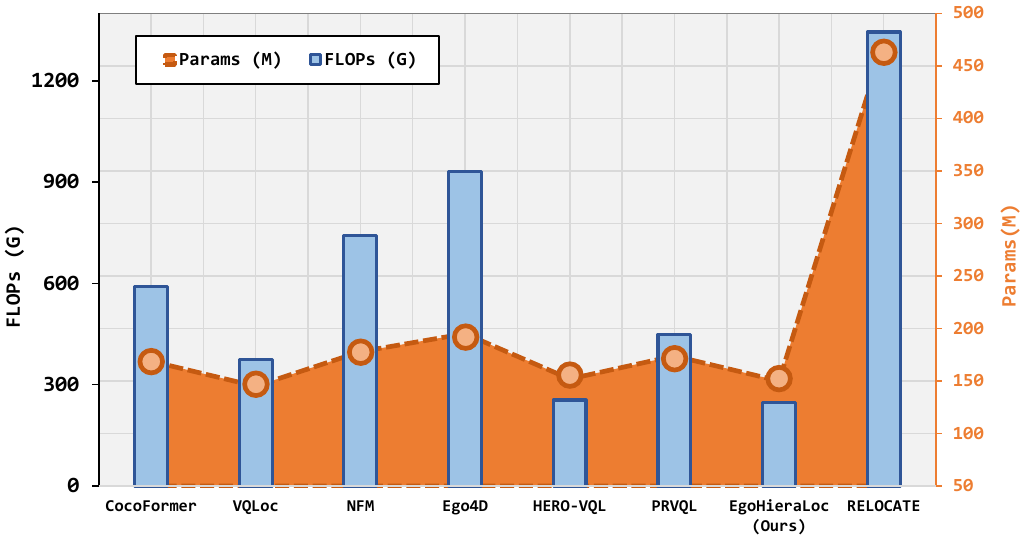}
    \caption{\textbf{Analysis of FLOPs and parameter scale.} Trade-off between computational cost in GFLOPs and model capacity in parameters, highlighting the structural efficiency of our architecture.}
    \label{fig:flopsparams}
  \end{minipage}%
  \hfill 
  \begin{minipage}[t]{0.49\textwidth}
    \centering
    \includegraphics[width=\linewidth]{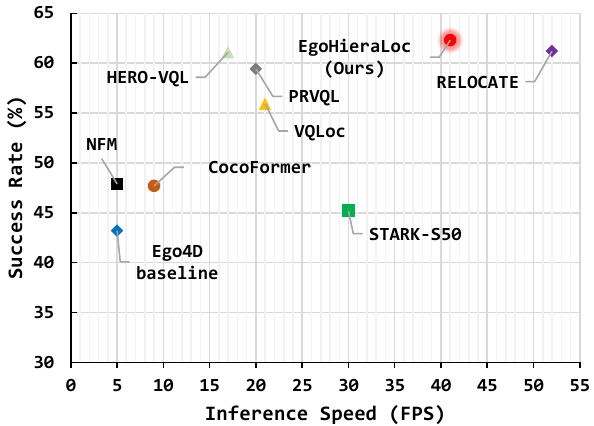}
    \caption{\textbf{Comparison of inference speed among current methods.} Inference speed comparison on the Ego4D-VQ2D task, where our method establishes an optimal Pareto frontier performance.}
    \label{fig:fps}
  \end{minipage}
\end{figure}
\section{Analysis of Computational Efficiency and Parameter Scale}
\label{app:flopsparams}
We evaluate the inference efficiency of our 2D branch on the VQ2D task to characterize the framework's operational footprint. As depicted in the accuracy-versus-speed trade-off curve (Fig.~\ref{fig:fps}), EgoHieraLoc establishes a new Pareto frontier, effectively reconciling localization precision with inference throughput. While established baselines such as Ego4D (5 FPS, 43.2\%) and NFM (5 FPS, 47.9\%) are severely constrained by low throughput, and tracking-based baselines like STARK-ST50 (30 FPS, 45.2\%) suffer from suboptimal accuracy, EgoHieraLoc achieves a state-of-the-art success rate of 62.3\% at a high-speed regime of 41 FPS. Although RELOCATE achieves a slightly higher throughput (52 FPS), it lags behind our framework in precision while incurring prohibitive computational costs. Compared to contemporary high-performance VQL frameworks such as HERO-VQL (17 FPS, 61.1\%), PRVQL (20 FPS, 59.4\%), and VQLoc (21 FPS, 55.9\%), EgoHieraLoc delivers over a 2$\times$ speedup while simultaneously outperforming them all in success rate. This validates that our hierarchical architecture mitigates the classical trade-off where fine-grained spatiotemporal reasoning necessitates prohibitive latency.

To further elucidate our structural efficiency, Fig.~\ref{fig:flops_params} details the joint trade-off between computational complexity (FLOPs) and model capacity (Parameters). EgoHieraLoc exhibits superior computational economy, requiring a lightweight 248G FLOPs per inference—achieving the lowest FLOPs among all benchmarked methods. This represents a drastic reduction in computational overhead relative to heavy architectures like Ego4D (933G), NFM (742G), and Cocoformer (590G). In terms of model capacity, EgoHieraLoc maintains a compact 150M parameter footprint. While VQLoc occupies a marginally smaller parameter size (145M), its computational demand (374G FLOPs) is 50.8\% higher, indicating lower capacity utilization efficiency. Conversely, although HERO-VQL matches our parameter scale (155M) with comparable FLOPs (255G), its operational speed is severely bottlenecked (17 FPS vs. 41 FPS) due to unoptimized temporal operations. Finally, the training-free RELOCATE requires 472M parameters and 1345G FLOPs—exceeding our model by 3.1$\times$ in parameters and 5.4$\times$ in FLOPs. Relative to the original Ego4D baseline (195M), our cortically inspired design achieves a 23\% parameter reduction while dramatically boosting both accuracy and throughput.

It is pertinent to note that our efficiency analysis focuses on the intrinsic computations of the model via the VQ2D branch. We deliberately exclude end-to-end VQ3D latency from the throughput analysis, as 3D localization is dominated by external camera pose estimation (e.g., COLMAP). The computational cost of these offline pre-processing components fluctuates drastically with scene geometry and video duration, which would introduce stochastic noise into the evaluation of the model's core algorithmic efficiency. Given that our 3D branch inherits the hierarchical backbone and feature representations of the 2D branch, this analysis provides a definitive and representative assessment of the framework's computational characteristics.
\section{Pseudo code of Core Modules}
\label{app:pseudocode}
We provide formal algorithmic specifications for the two core modules of 
EgoHieraLoc: the DPM and the QAM. These pseudocode listings complement the architectural 
descriptions in Section~\ref{sec:method} by making explicit the 
step-by-step computational procedures, hyperparameter dependencies, and 
data flow that underpin each module. Algorithm~\ref{code:dpm} details the 
DPM pipeline, which partitions query features into SAM-guided foreground 
and background token sets $\mathcal{M}_O$ and $\mathcal{M}_B$, computes 
per-location cosine similarity scores against both sets via Top-$K$ 
aggregation, and produces a foreground response map $\mathcal{O}$ and a 
likelihood-channel map $\mathcal{Z}$ for downstream fusion. 
Algorithm~\ref{code:qam} details the QAM pipeline, which trains a 
Discriminative Correlation Filter (DCF) directly on query features in the 
frequency domain, performs adaptive peak selection via connected-component 
analysis, and refines frame-level location estimates through a 
deformation-aware distance metric that jointly accounts for cumulative 
motion trajectory $\Omega^{(v_i)}$ and dynamically updated ellipse 
geometry $(a^{(v_i)}, b^{(v_i)})$. Together, these two modules constitute 
the spatial grounding backbone of EgoHieraLoc, whose outputs are fused in 
the RAM to produce the final per-frame 
localization response.
\begin{CJK*}{UTF8}{gkai}
\label{code:dpm}
\begin{algorithm}
    \caption{Discriminative Parsing Module (DPM)}
    \begin{algorithmic}[1]
        \Require Query $\mathcal{Q} \in \mathbb{R}^{h \times w \times c}$, Search region features $\mathcal{F}(v_i) \in \mathbb{R}^{H \times W \times D}$, Top-K parameter $K_{DPM}$
        \Ensure Foreground map $\mathcal{O}$, Likelihood-channel map $\mathcal{Z}$

        \State Segment $\mathcal{Q}$ via SAM to obtain binary mask $\mathcal{S}(\mathcal{Q}) \in \{0,1\}^{h \times w}$
        \State Extract query features $\mathcal{F}(\mathcal{Q}) \in \mathbb{R}^{h' \times w' \times d}$ via backbone
        \State Compute scaling ratio $s = \lfloor h / h' \rfloor$; align $\mathcal{F}(\mathcal{Q})$ with $\mathcal{S}(\mathcal{Q})$ via bilinear interpolation

        \State Partition query features into foreground and background sets:
        \Statex \quad $\mathcal{M}_O = \{ \mathcal{F}(\mathcal{Q})_{(x',y')} \mid \mathcal{S}(\mathcal{Q})_{(x's,\, y's)} = 1 \}$
        \Statex \quad $\mathcal{M}_B = \{ \mathcal{F}(\mathcal{Q})_{(x',y')} \mid \mathcal{S}(\mathcal{Q})_{(x's,\, y's)} = 0 \}$

        \State Process search region features through two CBN layers:
        \Statex \quad $\mathcal{F}(v_i)' = CBN_2(CBN_1(\mathcal{F}(v_i))) \in \mathbb{R}^{H_\xi \times W_\xi \times D_\xi}$

        \ForAll{spatial location $(x, y)$ in $\mathcal{F}(v_i)'$}
            \ForAll{$m_o \in \mathcal{M}_O$}
                \State $L_{(x,y),m_o}^O = \dfrac{\mathcal{F}(v_i)'_{(x,y)} \cdot m_o}{\max(\|\mathcal{F}(v_i)'_{(x,y)}\|_2 \cdot \|m_o\|_2,\; \epsilon)}$
            \EndFor
            \ForAll{$m_b \in \mathcal{M}_B$}
                \State $L_{(x,y),m_b}^B = \dfrac{\mathcal{F}(v_i)'_{(x,y)} \cdot m_b}{\max(\|\mathcal{F}(v_i)'_{(x,y)}\|_2 \cdot \|m_b\|_2,\; \epsilon)}$
            \EndFor
            \State $\mathcal{O}_{(x,y)} = \dfrac{1}{K_{DPM}} \displaystyle\sum_{L \in \mathrm{TopK}(L_{(x,y)}^O,\, K_{DPM})} L$
            \State $\mathcal{B}_{(x,y)} = \dfrac{1}{K_{DPM}} \displaystyle\sum_{L \in \mathrm{TopK}(L_{(x,y)}^B,\, K_{DPM})} L$
        \EndFor

        \State Compute likelihood-channel map: $\mathcal{Z} = \mathrm{softmax}([\mathcal{O},\, \mathcal{B}])$

        \Return $\mathcal{O}$, $\mathcal{Z}$
    \end{algorithmic}
\end{algorithm}
\end{CJK*}
\begin{CJK*}{UTF8}{gkai}
\label{code:qam}
\begin{algorithm}
    \caption{Query-Aware Module (QAM)}
    \begin{algorithmic}[1]
        \Require Query $\mathcal{Q}$, Search region frames $\{v_i\}$, sensitivity $\alpha_{QAM}$, decay $\gamma_{QAM} = 0.4$, smoothing $\eta_{QAM} = 0.7$, ellipse weight $\lambda_e = 0.5$
        \Ensure Refined location $p_{\text{mod}}^{(v_i)}$ and location response map for each frame $v_i$

        \State Train DCF filter $f$ on $\mathcal{F}(\mathcal{Q})$ by minimizing:
        \Statex \quad $\mathcal{L}_f = \operatorname*{argmin}_f \|\mathcal{FT}^{-1}(\mathcal{FT}(\mathcal{F}(\mathcal{Q})) \odot \overline{\mathcal{FT}(f)}) - y\|^2_2 + \lambda\|f\|^2_2$
        \State Initialize ellipse axes $(a^{(v_0)}, b^{(v_0)})$ from query; set $\Omega^{(v_0)} = 0$

        \ForAll{frame $v_i$}
            \State Compute response map: $R_{v_i} = \mathcal{FT}^{-1}(\mathcal{FT}(\mathcal{F}(v_i)) \odot \overline{\mathcal{FT}(f)})$
            \State Compute adaptive threshold: $T_{v_i} = \mu_{R_{v_i}} + \alpha_{QAM} \cdot \sigma_{R_{v_i}}$
            \State Apply threshold to get binary mask $M_{v_i}$; extract connected components $\{CC_k\}_{k=1}^N$ via 8-connectivity
            \ForAll{connected component $CC_k$}
                \State Extract candidate peak: $p_k^{(v_i)} = \operatorname*{argmax}_{(x,y) \in CC_k^{(v_i)}} R_{v_i}(x, y)$
            \EndFor
            \State Retain Top-5 peaks by response value; select primary candidate:
            \Statex \quad $p_{\text{can}}^{(v_i)} = \operatorname*{argmax}_{p_k^{(v_i)} \in \text{Top-5}} \left[ R_{v_i}(p_k^{(v_i)}) \cdot \mathcal{A}(CC_k^{(v_i)}) \right]$
            \If{$i = 0$}
                \State Set $\Delta^{(v_0)} = 0$; assign $p_{\text{mod}}^{(v_0)} = p_{\text{can}}^{(v_0)}$
            \Else
                \State Compute instantaneous deformation: $\Delta^{(v_i)} = p_{\text{can}}^{(v_i)} - p_{\text{mod}}^{(v_{i-1})}$
                \State Update cumulative deformation: $\Omega^{(v_i)} = \gamma_{QAM} \cdot \Omega^{(v_{i-1})} + (1 - \gamma_{QAM}) \cdot \Delta^{(v_i)}$
            \EndIf
            \State Compute bounding box of $CC_{\text{can}}^{(v_i)}$ to get $w^{(v_i)}$, $h^{(v_i)}$; update semi-axes:
            \Statex \quad $a^{(v_i)} = \eta_{QAM} \cdot a^{(v_{i-1})} + (1-\eta_{QAM}) \cdot w^{(v_i)} / 2$
            \Statex \quad $b^{(v_i)} = \eta_{QAM} \cdot b^{(v_{i-1})} + (1-\eta_{QAM}) \cdot h^{(v_i)} / 2$
            \ForAll{coordinate $i = (x_i, y_i)$ in $R_{v_i}$}
                \State Compute motion-aligned coordinate: $i' = i + \psi(\Omega^{(v_i)})$
                \State Compute deformation-aware distance:
                \Statex \quad $Dis_i^{\text{mod},(v_i)} = \|i' - p_{\text{can}}^{(v_i)}\|_2 + \lambda_e \cdot \sqrt{\dfrac{(x_i - p_x)^2}{(a^{(v_i)})^2} + \dfrac{(y_i - p_y)^2}{(b^{(v_i)})^2}}$
            \EndFor
            \State Determine refined location: $p_{\text{mod}}^{(v_i)} = \operatorname*{argmin}_i\, Dis_i^{\text{mod},(v_i)}$
        \EndFor

        \Return $p_{\text{mod}}^{(v_i)}$ and location response map for each $v_i$
    \end{algorithmic}
\end{algorithm}
\end{CJK*}

\end{document}